\documentclass{article}

\usepackage{arxiv}

\usepackage[utf8]{inputenc} 
\usepackage[T1]{fontenc}    
\usepackage[hidelinks]{hyperref}       
\usepackage{url}            
\usepackage{booktabs}       
\usepackage{amsfonts}       
\usepackage{nicefrac}       
\usepackage{microtype}      
\usepackage{lipsum}		
\usepackage{graphicx}
\usepackage{natbib}
\usepackage{doi}
\usepackage{csquotes}
\usepackage{acro}
\usepackage{placeins}
\usepackage{appendix}
\usepackage{float}

\DeclareAcronym{NLP}{short = NLP, long  = natural language processing}
\DeclareAcronym{ML}{short = ML, long  = machine learning}
\DeclareAcronym{LLM}{short = LLM, long  = Large Language Model}
\DeclareAcronym{LLMs}{short = LLMs, long  = Large Language Models}

\title{Positioning Political Texts with Large Language Models by Asking and Averaging}

\author{ \href{https://orcid.org/0000-0003-4800-0598}{\includegraphics[scale=0.06]{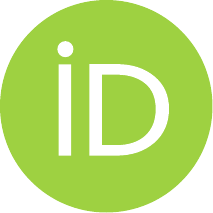}\hspace{1mm}Ga\"el Le~Mens}\thanks{This research was funded by ERC Consolidator Grant 772268 from the European Commission to G.L.M, ICREA Academia grants to A.G and G.L.M, grants PID2021-123111OB-I00 (A.G.) and PID2022-137908NB-I00 (G.L.M.) funded by MICIN/AEI/10.13039/501100011033 and by ERDF/UE ``A way of making Europe'', and the Severo Ochoa Programme for Centres of Excellence in R\&D (Barcelona School of Economics CEX2019-000915-S) funded by \\MCIN/AEI/10.13039/501100011033. 
We thank Hauke Licht, Christopher Wratil, Xavier Fernandez-Marin, Fabrizio Gilardi, Kenneth Benoit, and participants in the AI-PSR workshop at the University of Barcelona for valuable comments and discussion.
This article has  been conditionally accepted in \emph{Political Analysis}. Earlier titles: ``Scaling Political Texts with ChatGPT'' and ``Scaling Political Texts with Large Language Models: Asking a Chatbot Might Be All You Need''.} \\
	Department of Economics and Business \\
    Universitat Pompeu Fabra, \\ 
    Barcelona School of Economics, \\
    UPF-Barcelona School of Management, \\
    Barcelona, Spain \\
	\texttt{gael.le-mens@upf.edu} \\
	\And
	\href{https://orcid.org/0000-0001-7943-3500}{\includegraphics[scale=0.06]{orcid.pdf}\hspace{1mm}Aina Gallego} \\
	Department of Political Science \\
    Constitutional Law and Philosophy of Law \\
    Universitat de Barcelona,  \\
    Institut Barcelona d’Estudis Internacionals, \\
    Barcelona, Spain \\
	\texttt{aina.gallego@ub.edu} \\
}

\renewcommand{\shorttitle}{Positioning Political Texts with LLMs by Asking and Averaging}
\hypersetup{
pdftitle={Positioning Political Texts with LLMs by Asking and Averaging},
pdfsubject={LLM,  ideology, scaling, text as data},
pdfauthor={Ga\"el Le Mens},
pdfkeywords={Aina Gallego},
}

\begin{document}
\maketitle

\begin{abstract}
    We use instruction-tuned Large Language Models (LLMs) like GPT-4, Llama 3, MiXtral, or Aya to position political texts within policy and ideological spaces. We ask an LLM where a tweet or a sentence of a political text stands on the focal dimension and take the average of the LLM responses to position political actors such as US Senators, or longer texts such as UK party manifestos or EU policy speeches given in 10 different languages. The correlations between the position estimates obtained with the best LLMs and benchmarks based on text coding by experts, crowdworkers, or roll call votes exceed .90. This approach is generally more accurate than the positions obtained with supervised classifiers trained on large amounts of research data. Using instruction-tuned LLMs to position texts in policy and ideological spaces is fast, cost-efficient, reliable, and reproducible (in the case of open LLMs) even if the texts are short and written in different languages. We conclude with cautionary notes about the need for empirical validation. 
\end{abstract}


\section{Introduction}
Much research in the social and political sciences involves estimating the positions of actors, such as politicians or political parties, in latent ideological and policy spaces, such as the left-right or liberal-to-conservative continuum. Widely used approaches involve the automatic processing of text documents produced by these actors, such as party manifestos \citep{ laver2003extracting, slapin2008scaling} or legislative speeches \citep{lauderdale2016measuring}. Other approaches involve human coding by experts \citep{budge2001mapping} 
or crowd workers \citep{benoit2016crowd}. Yet other approaches rely on other inputs, such as roll call votes \citep{poole1985spatial}, Twitter connections \citep{barbera2015birds}, or campaign donations \citep{bonica2014mapping}.

We propose a new approach to position text documents in ideological and policy spaces using instruction-tuned Large Language Models (LLMs) and evaluate its performance. These models are LLMs optimized for dialogue use cases and are typically interacted with via chatbots such as ChatGPT. We build on the direct query method introduced by \cite{LeMens2023_PNASGPT4_Typicality} who measured the typicality of text documents in concepts by asking GPT-4 for typicality scores. We directly ask an LLM where a tweet or a sentence of a political text stands on the focal dimension and take the average of the LLM responses to obtain position estimates of longer texts or political actors.

We focus on four scaling tasks using texts of different types, contexts, and lengths. First, we position individual tweets published by US Representatives and Senators by directly asking LLMs where these stand on the left-right ideological spectrum.
Second, we position senators of the 117th US Congress on the same dimension by averaging the position estimates of a sample of the tweets they published during the Congress session. 
Third, we position party manifestos on the economic and social policy dimensions. We ask the LLMs for the positions of each sentence on these dimensions and average the LLM responses to obtain position estimates of the party manifestos. Fourth, we apply this approach to position speeches by EU legislators, in 10 different languages, about a policy proposal on the `anti-subsidy’ to ‘pro-subsidy’ scale. This allows us to explore the potential of this approach for comparative research with multilingual data. 

Several articles have shown that LLMs can produce text \textit{annotations} (classifications in discrete categories such as relevant/irrelevant or a topic among a limited set of candidate topics) that are in good agreement with those produced by human coders \citep[e.g.,][]{Gilardi_chatGPT_1, tornberg2023chatgpt, ziems2023can}. However, little work has used LLMs to produce position estimates of political texts in ideological and policy spaces, which is the essence of text \textit{scaling}, a core task in political science (see \cite{benoit2020text} for a discussion of the difference between the two tasks). We know of only two recent studies that rely on the text generation capabilities of LLMs to estimate policy and ideological positions. Our approach differs from both. 
The first study used GPT-3 as a probabilistic text classifier to obtain the posterior probability that a sentence in a party manifesto is ``Conservative'' or ``Liberal'' and defined the position of the manifesto as the average (across sentences) of the difference between these probabilities \citep{ornstein2022train}. Our approach directly asks the LLM for the position of text on the focal dimension, and we find that it performs better.\footnote{The correlations between the Expert coding estimates and the positions produced by \cite{ornstein2022train} are $.92$ and $.8$. See Figures~\ref{fig_result_UK_manifestos_economic} and~\ref{fig_result_UK_manifestos_social} for our results.} 
The second study asked GPT 3.5 and Llama 2 to compare pairs of politicians on a particular dimension and used these pairwise comparisons to construct estimates of politician position on unidimensional scales (e.g., gun control support) \citep{wularge}. We ask LLMs to position political \emph{texts} produced by political actors instead of asking them to position political actors based on their names. Our approach is thus applicable even to political actors about whom the LLM has little information.

\section{Methods and data}
\subsection{Obtaining position estimates with LLMs}

Table~\ref{tab_LLMs_info_body} lists the LLMs we used for text scaling, including their open or closed status. These consist of a set of the most recent and largest LLMs available at the end of May 2024.\footnote{See table~\ref{tab_USCongress_118_body_nb_responses_appendix} for analyses with other model such as Mistral, Gemma or Llama 2.}

 \begin{table}[htbp]
    \centering
    \includegraphics[width = \columnwidth]{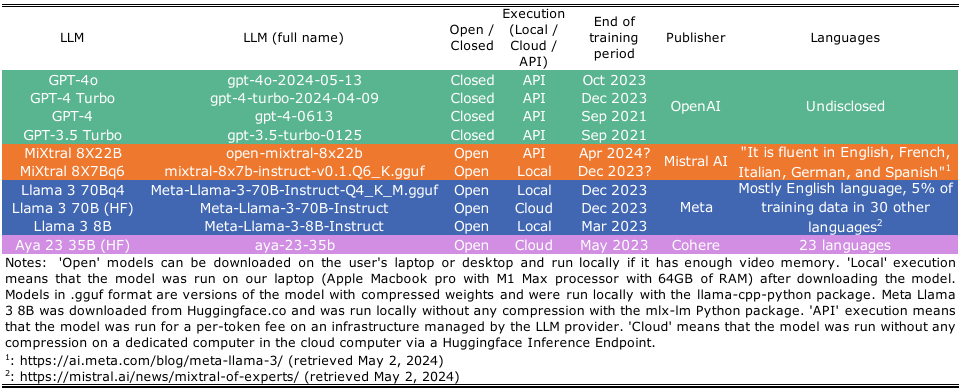}
    \caption{LLMs used for the comparative analyses. 
    }    \label{tab_LLMs_info_body}
    \vspace{0.1cm}
\end{table}

To obtain a position estimate of a text with an LLM, we submitted a prompt that contained a `user message' instructing it to return such an estimate. For example, to locate a tweet on the left-to-right-wing scale, we used:

\begin{displayquote}\small
    You will be provided with the text of a tweet published by a member of the US Congress. Where does this text stand on the `left' to `right' wing scale?  Provide your response as a score between 0 and 100 where 0 means `Extremely left' and 100 means `Extremely right'. If the text does not have political content, set the score to "NA". You will only respond with a JSON object with the key Score. Do not provide explanations.

    $\ll$  Text of the tweet $\gg$ 
\end{displayquote}

In all cases, we set the temperature parameter to 0, to ensure that the LLM would generate its response by selecting the most likely next token, and thus make the LLM responses as deterministic as possible (to ensure replicability). We also set the maximum number of tokens in the response to 20. This parameter does not affect the nature of the message returned by LLMs; it cuts the response down to 20 tokens if the LLM intended to generate a longer response. This ensures speed (token generation tends to be relatively slow) and limits costs (pay-per-use APIs charge per token submitted in the prompt and per token returned in the response). Finally, whenever this option was available, we set the response format to be a JSON object.  

To obtain the position of a party manifesto or a policy speech, we proceeded in a similar way with each sentence of the text documents. We then took the average of the positions of the sentences for which the LLM returned a numeric score, mimicking the approach used by \cite{benoit2016crowd} with human coders. 


\subsection{Data}

\subsubsection{Tweets published by US Congress members after the training cut-off date of GPT-4}
These data allow us to assess the performance of a modern LLM on prediction data we are certain were not part of the LLM pre-training data. We used the 900 tweets originally analyzed in \cite{LeMens2023_PNASGPT4_Typicality}. In November 2023, we recruited 597 Prolific participants to each rate 30 tweets by answering the following question: ``Where does this text stand on the `left' to `right' wing scale? If the text does not have political content, select `Not Applicable'.'' Participants were not given any instructions as to what we mean by `left' or `right.' The crowdsourced position estimate of a tweet is the average of these ratings. All tweets, except one, received at least one position rating (different from `NA'), leading to a test data set of 899 tweets and their crowdsourced position estimates. This measure is highly reliable overall and within-party (table~\ref{tab_benchmark_reliability}).

\begin{table}[htbp]
    \centering
    
    \includegraphics[width = .5\columnwidth]{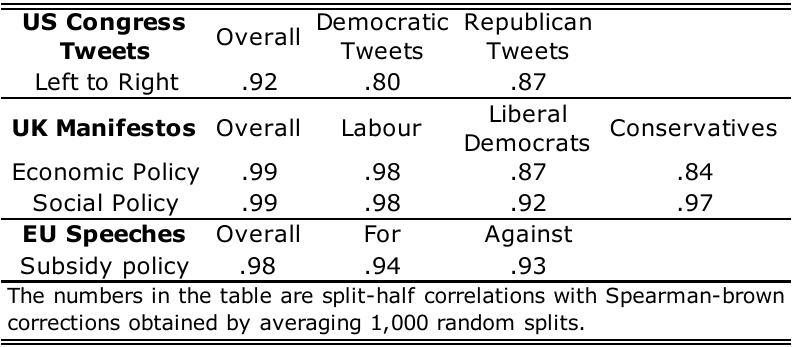}
    \caption{Reliability of the measures based on human ratings used as benchmark for assessing the performance of position estimates produced with LLMs. 
    }    \label{tab_benchmark_reliability}
    \vspace{0.1cm}

\end{table}

\subsubsection{Senators of the 117th Congress}
We obtained the list of senators from VoteView.com, their Twitter usernames, and downloaded the tweets they published during the Congress session through the Twitter API. We used random samples of 100 tweets published by each senator during the 117th Congress session (January 3, 2021 to January 3, 2023).
We excluded two senators who published fewer than 100 tweets during the Congress sessions. We use as a benchmark the first dimension Nokken-Poole period-specific DW-NOMINATE score, a well-established position estimate based on senators' roll-call votes.

\subsubsection{British party manifestos}
We obtained the texts, expert coding estimates, and crowd coding estimates from the replication package of \cite{benoit2016crowd}. We positioned the 18 British party manifestos on an economic policy dimension (from left- to right-wing) and on a social policy dimension (from conservative to liberal). We used as a benchmark the Expert Coding estimates that were constructed by \citeauthor{benoit2016crowd} based on the sentence position estimates provided by a crowd of experts (political scientists). This measure is overall highly reliable and has varying levels of within-party reliability (table~\ref{tab_benchmark_reliability}).

\subsubsection{Multilingual setting: EU policy speeches in 10 languages}
We also positioned the 36 speeches of a European Parliament debate on a policy proposal concerning state subsidies originally analyzed in \citeauthor{benoit2016crowd} on the pro- to anti-subsidy dimension. These were delivered in 10 different languages by speakers who then voted for or against the proposal. \cite{benoit2016crowd} obtained 6 crowdsourced position estimates for each speech from crowdworkers coding the official translations in English, German, Greek, Italian, Polish, and Spanish. We took the simple average of the six crowd-coding estimates as a benchmark. This setting is challenging not only because of its multilingual nature but also because the speeches vary in style (e.g. technical, case-focused, rhetorical) and require knowledge of the debate context to be understood. 

\section{Results}

\subsection{Tweets published by members of the US Congress after the training cut-off date of GPT-4} \label{sub_results_congress_118_crowdsourcing}

\begin{figure}[htbp]
    \centering
    \includegraphics[width = \columnwidth]{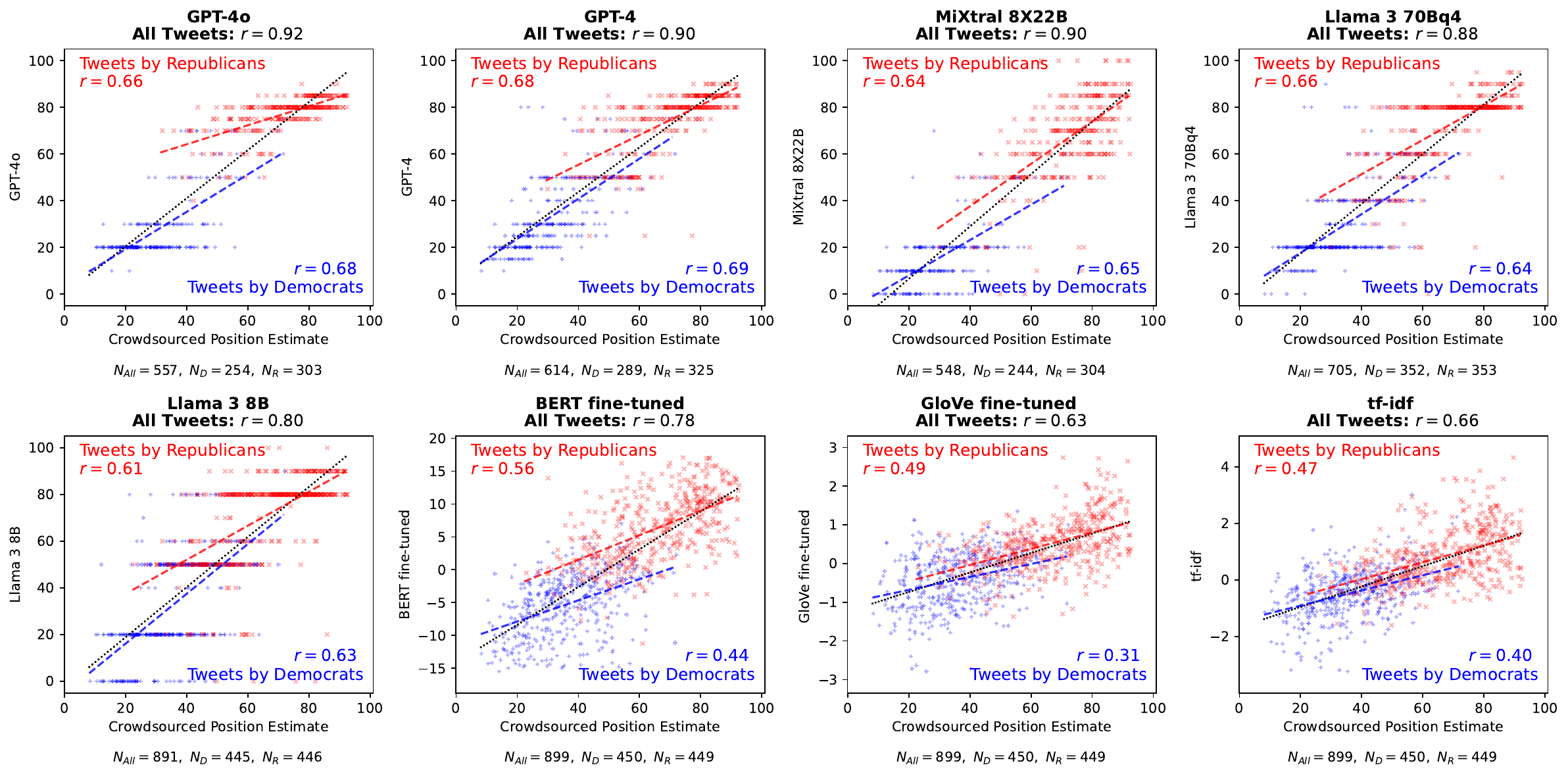}
    \caption{Positioning tweets published by members of the US Congress on the left-right ideological spectrum ($N=899$). 
    }    \label{fig_tweets_after_cutoff_baseline}
    \vspace{0.1cm}

\end{figure}

The LLMs returned `NA' for a subset of tweets, indicating that they judged that these tweets did not have enough political content to return a position estimate (see~\ref{sec_additional_results_tweets_congress_118} for further discussion).
The correlations between the position estimates produced by the best LLMs and crowdsourcing are very high, as shown in Figure~\ref{fig_tweets_after_cutoff_baseline}. Position estimates reflect differences between-party and within-party. 

To compare these results with those obtained through approaches that do not require the submission of prompts to an LLM, we computed the typicality of each tweet in the Republican and the Democratic parties using probabilistic text classifiers and defined the position of a tweet as the difference between these two typicalities. The training data consist of approximately 1 million tweets published by members of the US Congress during the 116th and 117th Congress sessions. 

We used text classifiers based on fine-tuned BERT (the highest performing approach in \cite{LeMens2023SociologicalScience_BERT_typicality}), fine-tuned GloVe word embeddings and a naive Bayes classifier based on word frequencies (TF-IDF).\footnote{See Appendix~\ref{App_delta_typicality_LLMs} for an approach that asks LLMs to return typicality ratings.} None of these approaches matches the best-performing LLMs, especially when it comes to capturing within-party differences.

\subsection{Senators of the 117th US Congress}\label{sub_results_congress_117_senators}
This setting differs from the  previous one in that the benchmark positions are not based on human coding but on the voting \emph{behavior} of the senators. 

\begin{figure}[htbp]
    \centering
    \includegraphics[width = \columnwidth]{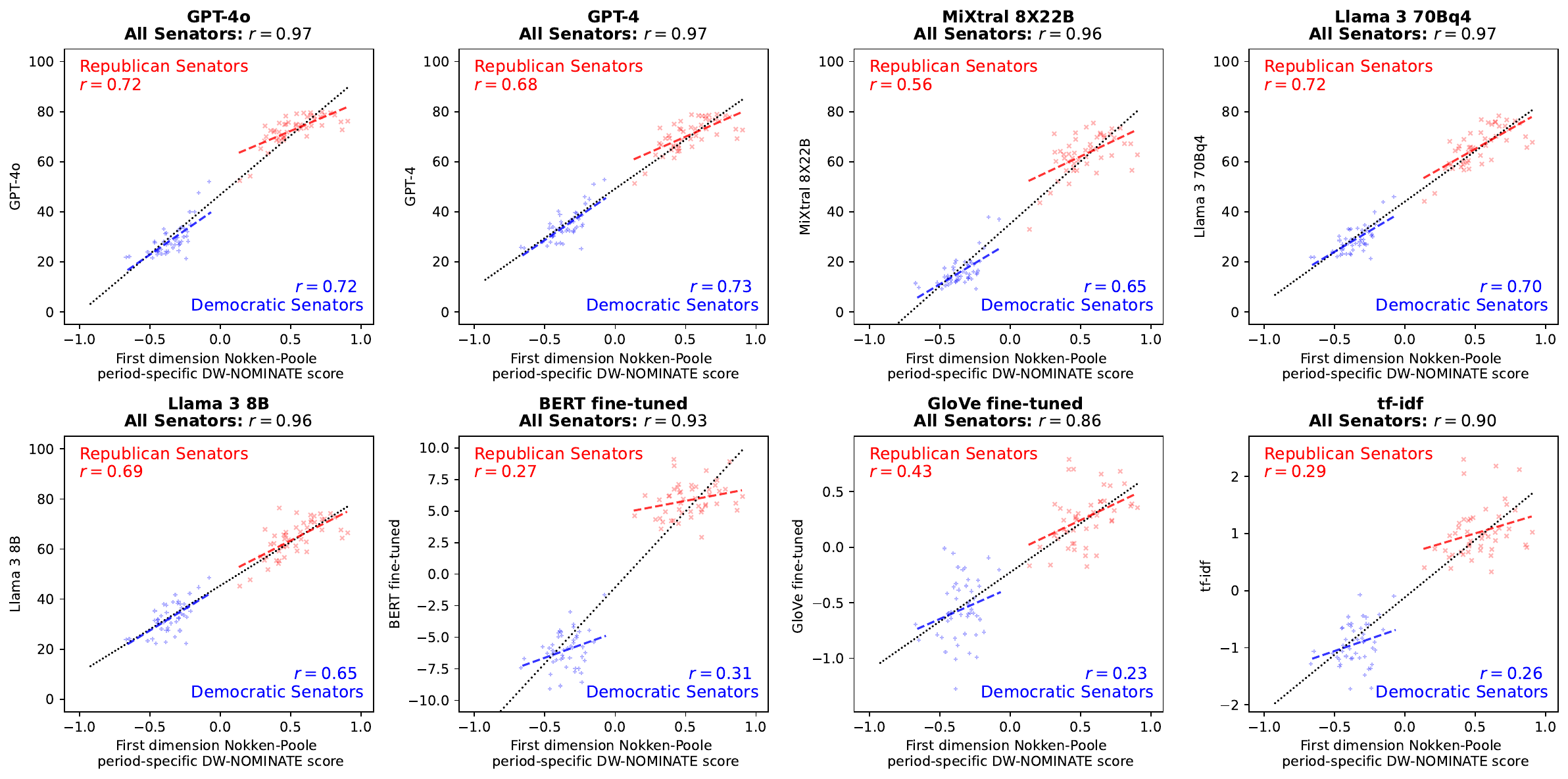}    
    \caption{Positioning Senators of the 117th Congress on the left-right ideological spectrum based on a random sample of 100 of their tweets ($N=98$). Each dot represents a senator (`{\sf +}': Democrats, `{\sf x}': Republicans, `{\sf o}': others). 
    }    \label{fig_senator_117_sentence_scaling_Nokken_Poole}
    \vspace{0.1cm}
\end{figure}

The position estimate of a senator is the average position of their tweets on the left-right ideological spectrum. Figure~\ref{fig_senator_117_sentence_scaling_Nokken_Poole} shows that the resulting position estimates are highly correlated with those based on the roll call votes of the senators during that Congress session \citep{nokken2004congressional}, overall and within the party. 
These correlations are also higher than those produced by supervised classifiers used to obtain typicality measures of tweets in the two parties. 
The position estimates produced with LLMs are also highly correlated with those based on campaign funding (2020 CF scores, \cite{bonica2014mapping}), although less so within-party (Figure~\ref{fig_senator_117_sets_of_tweets_CF_score_dyn}). 

\FloatBarrier

\subsection{British party manifestos}

\FloatBarrier

 \begin{figure}[htbp]
    \centering
    \includegraphics[width = \columnwidth]{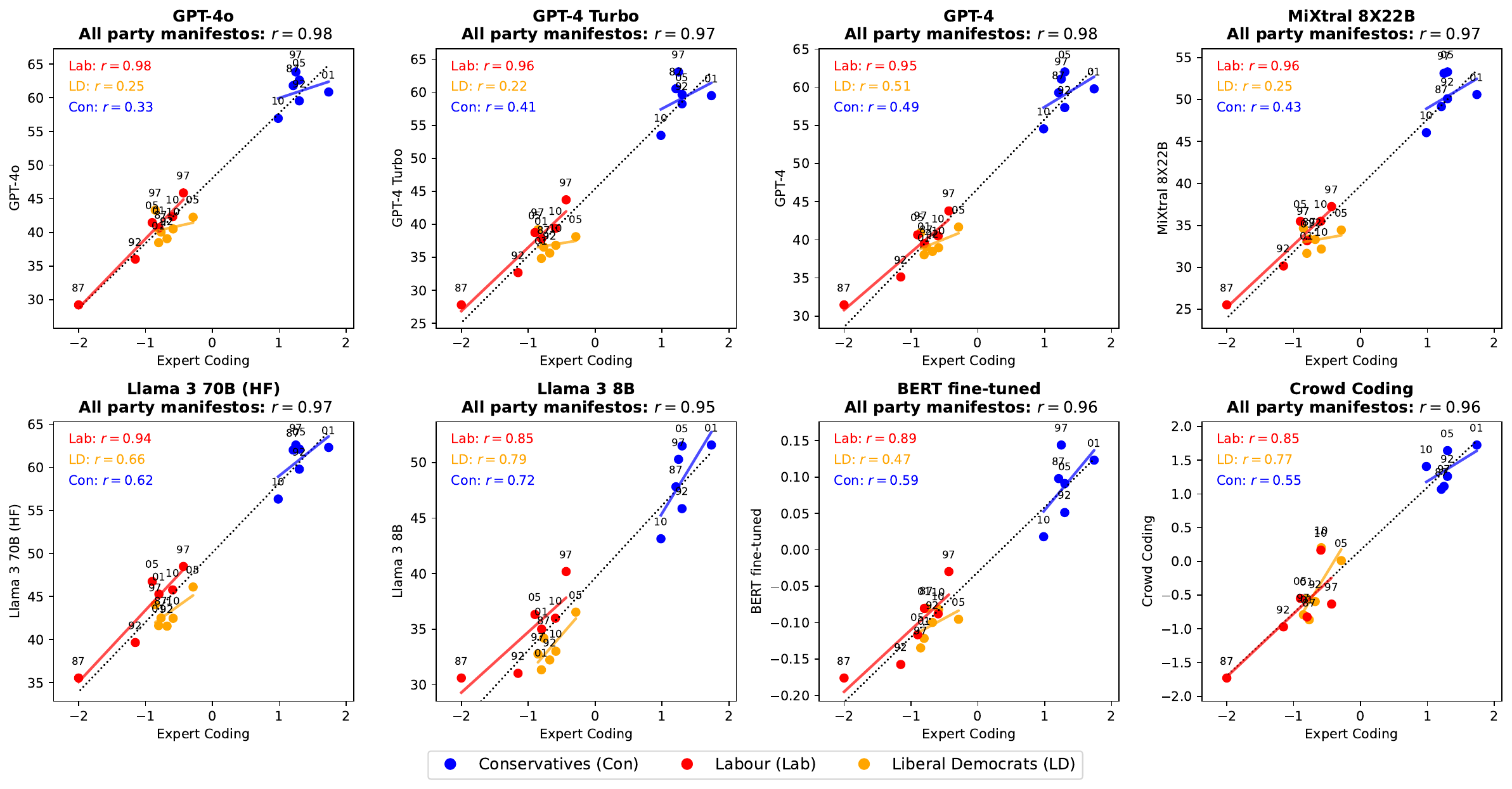}  
    \caption{Positioning British party manifestos on the Economic policy dimension (left to right wing scale). The numbers next to the dots indicate the years of the manifestos.  
    }    \label{fig_result_UK_manifestos_economic}
    \vspace{0.1cm}

\end{figure}

 \begin{figure}[htbp]
    \centering
    \includegraphics[width = \columnwidth]{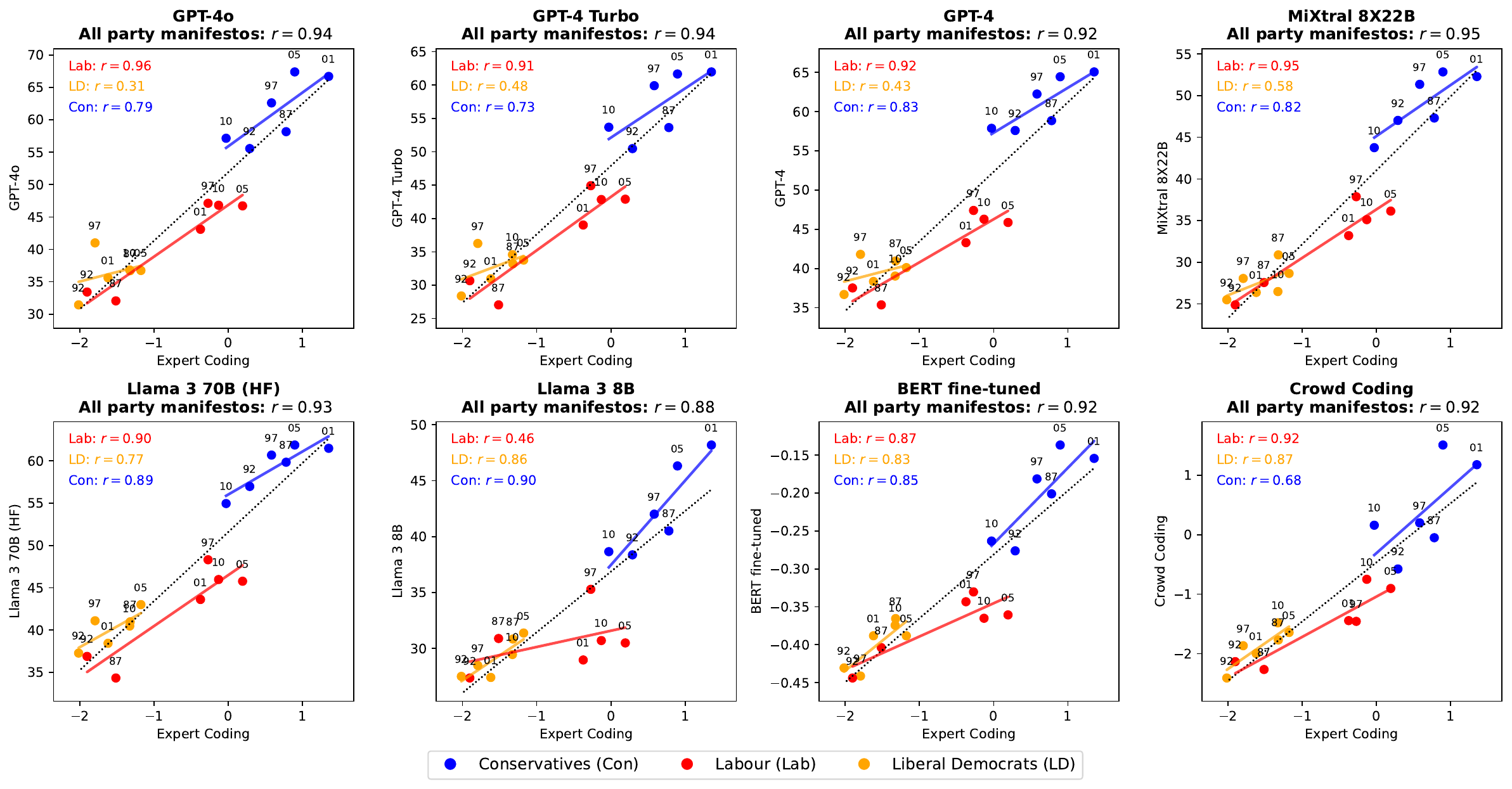}
    \caption{Positioning British party manifestos on the Social policy dimension (liberal to conservative scale). The numbers next to the dots indicate the years of the manifestos.
    }    \label{fig_result_UK_manifestos_social}
    \vspace{0.1cm}

\end{figure}

The position estimates obtained with the highest performing LLMs are very highly correlated with the Expert Coding estimates, at a level comparable to the position estimates produced with crowd workers (Figures~\ref{fig_result_UK_manifestos_economic} and ~\ref{fig_result_UK_manifestos_social}). This is the case not only overall, but also within political parties. These results were obtained without providing any description of the policy dimension to the LLMs. Similar results hold when including such descriptions (Figures~\ref{fig_result_UK_manifestos_economic_dim_desc} and ~\ref{fig_result_UK_manifestos_social_dim_desc}).

We also trained a BERT-based supervised probabilistic text classifier \citep{Devlin2018} using the crowdworkers' ratings collected by \cite{benoit2016crowd}, and used it to obtain position estimates of the manifestos' sentences and, in turn, of the manifestos. 
This approach did not yield better results than those obtained with the best LLMs, although the latter were (most likely) not specifically trained to position these party manifestos.

\FloatBarrier

\subsection{Multilingual setting: EU policy speeches in 10 Languages}

We obtained position estimates of the speeches on the `anti-subsidy’ to ‘pro-subsidy’ dimension by submitting each sentence to the LLMs \emph{in its original language} with instructions (in English) including background information on the context of the debate. 

\begin{figure}[htbp]
    \centering
    \includegraphics[width = \columnwidth]{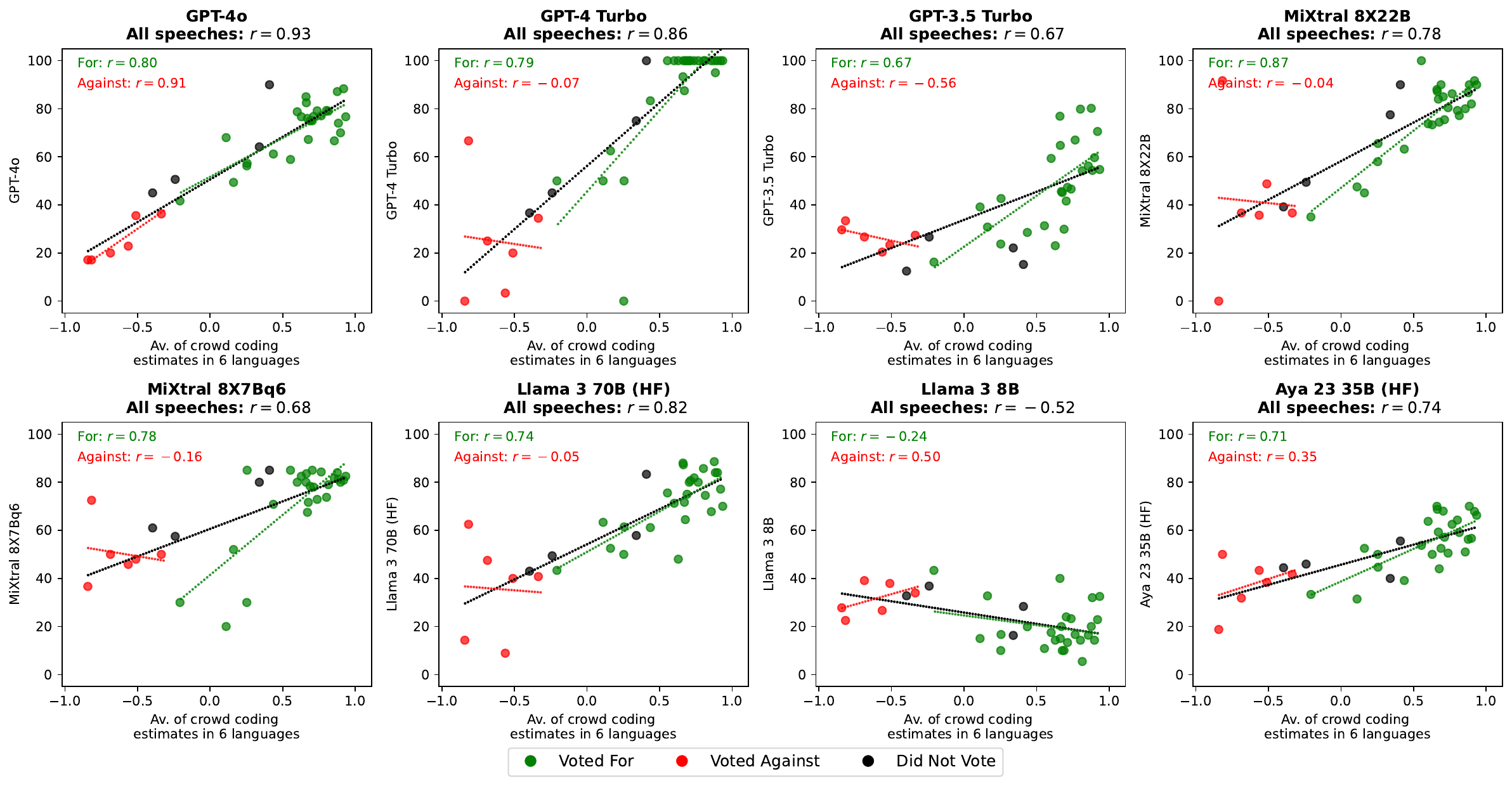}
    \caption{Positioning EU legislative speeches in 10 languages on the `anti-subsidy’ to `pro-subsidy’ dimension.
    } \label{fig_result_EU_speeches}   
    \vspace{0.1cm}
\end{figure}

For the highest performing LLMs (GPT-4o and, to a lesser extent, GPT-4 Turbo, MiXtral 8X22B, Llama 3 70B, Aya 23 35B), the correlation between the benchmark and the position estimates obtained is high overall, and also when we separate the speeches by speakers who voted for and against the policy (Figure~\ref{fig_result_EU_speeches}). 

Results obtained with the translations of the speeches in the 6 languages used to obtain the crowd coding estimates show that the best models perform well across languages  
(Appendix~\ref{app_EU_speeches_translations}).

\section{Discussion}

These results demonstrate that ``asking'' modern instruction-tuned LLMs for the position of short texts in ideological spaces can produce valid position estimates. These can be used, in turn, to position political actors such as politicians, as we illustrated with US senators. We showed that asking LLMs for the positions of sentences in ideological and policy spaces and averaging the responses produces valid positions of party manifestos and policy speeches. The position estimates of the party manifestos produced with the best LLMs are as accurate as the crowdsourced position estimates. And with individual tweets, ancillary analyses show that the position estimates returned by the best LLMs are as accurate as the average of the ratings of about 4 or more independent human coders (Appendix~\ref{app_sec_ENO_scores}).

This approach has the potential to expand the scope of text analysis due to its high accuracy, speed, ease of implementation, and reproducibility (for open LLMs). Moreover, querying instruction-tuned LLMs is much less costly than human coding even with the most expensive pay-per-use API (GPT-4): \$1.5 versus £1,626 for the 900 tweets analyzed in Section~\ref{sub_results_congress_118_crowdsourcing}.

Which LLM should researchers choose? Five main considerations come into play: accuracy with respect to a relevant benchmark, cost, speed, data protection, and reproducibility. If testing reveals no significant accuracy difference between open and closed LLMs, we recommend that researchers use open LLMs such as MiXtral (8X22B) and Llama 3 for text scaling tasks, at least in English. Their architecture and weights are freely available for download, which guarantees the reproducibility of research results and a high level of data protection (if executed locally or on a secure cloud machine). 

For multilingual settings, it seems that GPT-4o (a proprietary model) has a marked advantage compared to the best open models at the time of writing, but some LLM designers are releasing open LLMs especially designed to perform well in multiple languages (e.g., the Aya series by Cohere). Some LLMs perform better in some languages than in others (Appendix~\ref{app_EU_speeches_translations}) and systematic accuracy differences across languages can bias the results of downstream econometric analyses. Developing approaches to deal with this differential measurement error would help realize the potential of LLMs in comparative research.

Another decision is whether to position long text documents in a single prompt or to split them into shorter parts, such as sentences. At this stage, we do not have a clear recommendation on this issue. Positioning party manifestos in a single prompt leads to lower performance than the sentence-by-sentence and averaging approach (Appendix~\ref{app_manifestos_one_prompt}). In contrast, positioning senators by submitting their tweets in a single prompt did not cause significant performance degradation (Appendix~\ref{app_senators_one_prompt}). Assessing where and when the single-prompt approach leads to performance degradation is an interesting avenue for future research.

When interpreting the results of the LLM-based approach for positioning political actors such as politicians or parties, it is important to remember that position estimates are based on the text documents submitted to the LLMs. Therefore, the validity of the resulting estimates is limited by the information contained in these texts. In the case of party positions, our approach resembles the approach of the Comparative Manifesto Project but differs from approaches that rely on surveys of experts about their perception of party positions, such as the Chapel Hill Expert Survey. This also implies that the results obtained with LLMs might differ from those obtained from expert surveys in the same way that those obtained by human coding can differ from those obtained with expert surveys because the inputs used to produce the position estimates differ. 

The high correlations between position estimates produced with LLMs and human coders reported in this research note could tempt readers to use this approach in other domains while skipping the validation stage. But we advise them \emph{against doing so}. LLMs are well-known for generating biased and unreliable results in some empirical settings. Until other researchers have shown that asking (and averaging) a particular LLM provides accurate scaling results in a variety of empirical settings and focal latent dimensions, we cannot be sure about the breadth of settings in which LLMs perform well for scaling tasks and, \emph{a fortiori}, other measurement or coding tasks. Until more is known about the range of domains in which LLMs perform well at scaling and other measurement tasks, case-by-case empirical validation remains essential.





\bibliographystyle{chicago}

\bibliography{scalingWithLLMS} 

\vfill
\pagebreak

\appendix

\section*{\centering Appendix}
\bigskip

Prepared for "Positioning Political Texts with Large Language Models by Asking and Averaging" by Gaël Le Mens and Aina Gallego

\setcounter{figure}{0}
\renewcommand{\thefigure}{A\arabic{figure}}

\setcounter{table}{0}
\renewcommand{\thetable}{A\arabic{table}}
\section{Data}

\subsection{Tweets published by members of the US Congress after the training cut-off date of GPT-4 -- obtaining crowdsourced position estimates}

In November 2023, 597 Prolific participants (US residents, average age 40 years, 52\% male, 46\% female, 2\% others) each rated 30 tweets. We used the same 30 sets of 30 tweets, composed of 15 tweets from each party as in \cite{LeMens2023_PNASGPT4_Typicality}. Participants were randomly assigned to one of these sets. We obtained between 19 and 21 position ratings for each tweet, with an average of 20 ratings per tweet. The crowdsourced position estimate of a tweet is the average of these ratings. All tweets, except one, received at least one position rating (different from `NA'), leading to a test data set of 899 tweets and their crowdsourced position estimates.

To provide their position ratings, participants responded to the question: ``Where does this text stand on the `left' to `right' wing scale? If the text does not have political content, select `Not Applicable'.'' Participants were not given any instructions as to what we mean by `left' or `right.' They reported their scores using a slider with endings `Extremely Left' (left side, coded 0) to `Extremely right' (right side, coded 100). The slider was centered when the page appeared on the screen. Participants could use a checkbox to respond `Not Applicable'.

Importantly, participants were not provided with any information about the author of the tweet --- just the text of the tweet. Therefore, they were not informed whether the tweet was written by a Democrat or a Republican. 

To ensure response quality, we screened Prolific participants based on the following criteria:
\begin{itemize}
\item Fluent languages include English; 
\item Approval Rate: Minimum 95\%;
\item Country of Residence: US;
\item Political Spectrum (US): they should have provided a response to this question (the response could be anyone of \{Conservative, Moderate, Liberal, Other\}),
\item U.S. Political Affiliation: they should have provided a response to this question (the response could be anyone of \{Democrat, Republican, Independent, Other, None\}).
\end{itemize}
The last two items were included to increase the likelihood that participants would have some knowledge of US politics.
\noindent{}After providing informed consent, participants were asked to reflect on left- and right-wing tweets. They answered one question about each side of the ideological spectrum: 
\begin{displayquote}
What do you expect to see in a tweet that expresses a `left' [`right'] wing ideology? You could write about possible topics addressed by such tweets, and or opinions you expect the authors to have about these topics (min. response length: 100 characters). 
\end{displayquote}
The display order of the two questions was randomized among the participants. The responses to these questions were not analyzed (the purpose of these questions was to make participants think about the left-to-right ideological spectrum before they would respond to the scaling questions.

On the next page, they received short instructions about the position rating task. 

\begin{displayquote}
For each of the 30 tweets, you will be asked the following question:

Where does this tweet stand on the `left' to `right' wing scale? You will report your response using a continuous slider that goes from `Extremely left' to `Extremely right'. There is no right or wrong answer. We are interested in your subjective opinion.
If the text does not have political content, select `Not Applicable'. Note that the slider will appear on the screen 6 seconds after the text of the tweet.
\end{displayquote}

\noindent{}Then they looped through the 30 tweets and provided their position ratings for each of them.

Finally, the study ended with a short demographic questionnaire and some questions about their political orientation, their political opinions, and their frequency of use of Twitter (not analyzed in this article).

 \begin{figure}[htbp]
    \centering
    \includegraphics[width = .45\columnwidth]{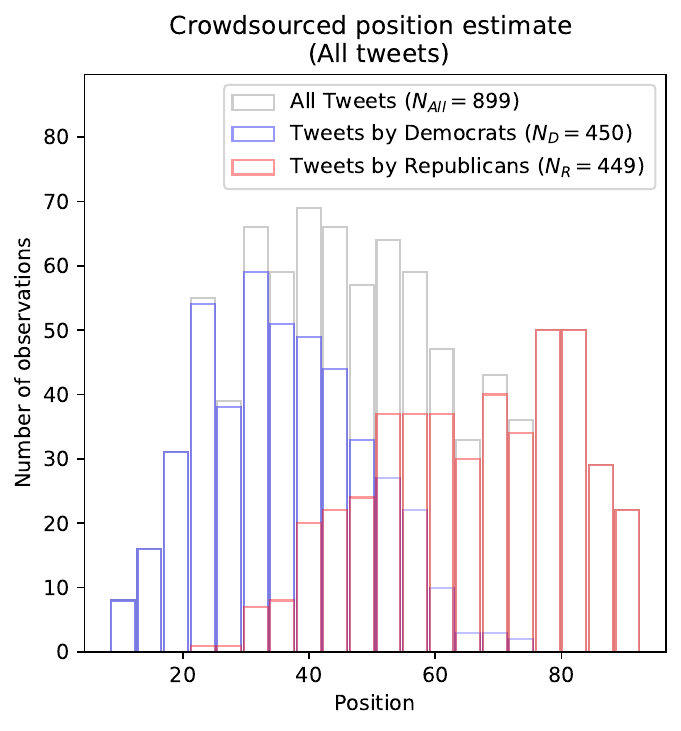}
    \hspace{.04\columnwidth}
    \includegraphics[width = .468\columnwidth]{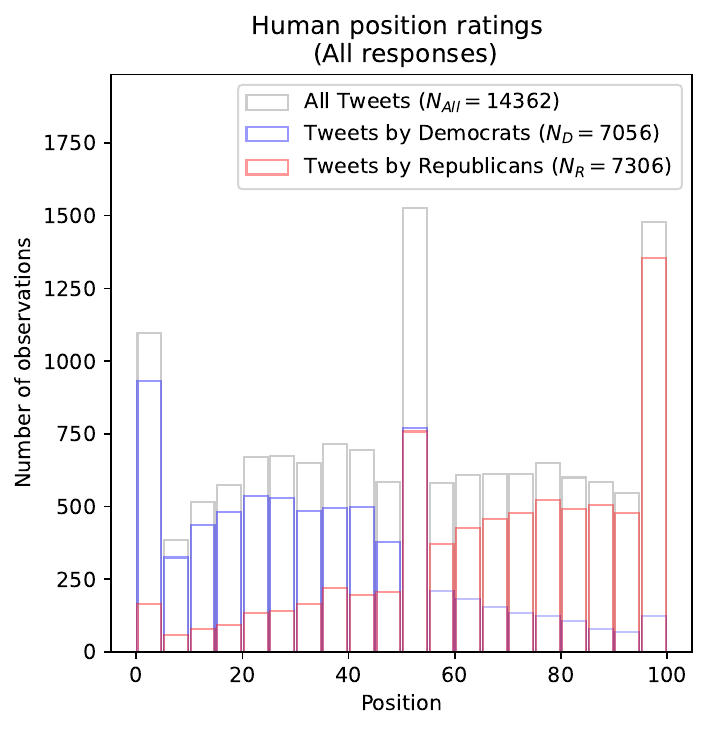}
    \caption{Tweets published by members of the 118th US Congress: Left panel:Histograms of responses by human respondents to the question about the position of each tweet on the Left-Right ideological spectrum (left panel). Right panel: Histogram of crowdsourced position estimates.
    }    \label{congress_118_tweets_histograms_crowdsourced_data}
    \vspace{0.1cm}

\end{figure}

\subsection{Senators of the 117th US Congress}
We obtained the list of senators from VoteView.com and the Twitter usernames of most senators from the data published with \cite{LeMens2023_PNASGPT4_Typicality} (\texttt{lattest\_legislators\_info.csv}). We complemented the list of usernames with a manual search on Twitter. We deposited a table that provides the Twitter usernames we used for each of the senators in the replication folder. (Twitter developer terms prevent us from posting the Tweets we used to position the senators but allow us to post the Tweet IDs of the tweets so that users with access to the API can easily re-download them).

We downloaded the tweets they published during the Congress session via the Twitter API (in 05/2023 and 11/2023). 

For each senator, we randomly sampled 100 tweets (with publication dates between January 3 and January 3, 2023). Those who published fewer than 100 tweets during the congressional session were excluded from the analysis, resulting in a set of 97 senators.

\subsection{British party manifestos}
We downloaded the replication materials for \cite{benoit2016crowd} from \href{
https://github.com/kbenoit/CSTA-APSR/tree/master}{https://github.com/kbenoit/CSTA-APSR/tree/master} 
(retrieved September 22, 2023). These consist of a folder \texttt{CSTA-APSR-master} from which we obtained the text of the party manifestos and the position estimates provided by expert surveys, sentence scaling by experts, and crowdsourced sentence scaling. 

The texts of the party manifestos used for scaling on the economic and social dimension were obtained from the file \texttt{CSTA-APSR-master/DataCreated/master.sentences.csv} which consisted of a table of sentences. 

We used the same procedure for the texts of the party manifestos used for scaling on the immigration dimension from the file \texttt{CSTA-APSR-master/DataCF jobs/immigration/}

\noindent{}\texttt{CFjobsresultsImmigration/f354277\_immigration1.csv}.

\subsection{EU legislative speeches}

We used the text of the speeches available in the replication package for \cite{benoit2016crowd} (in folder \texttt{07 Additional Texts}). To construct the crowdsourced position estimates of the speech based on translations in English, German, Greek, Italian, Polish, and Spanish, we used the raw crowdsourcing coding files (\texttt{fEnglishCombined.csv}, \texttt{f655316.csv}, \texttt{f656054.csv}, \texttt{f656640.csv}, \texttt{f658174.csv}, \texttt{f658191.csv}).

\newpage
\section{Constructing and submitting prompts to LLMs} 

We used different functions to interact with the different models (see below). Most of these functions use the same parameters and parameter values: 

\begin{itemize}
    \item \texttt{messages}: a Python list of dictionaries (with just one 1 element in our case) that contains the prompt (see details below).
    \item \texttt{max\_tokens}: 20. This parameter cuts the response of the LLM to a maximum of 20 tokens. It does not affect the nature of the response (it is not an instruction to respond in a few tokens). We used this small value to limit cost (pay-per-use APIs charge per number of tokens in the prompt and the response) and favor speed (the token generation process is relatively slow).
    \item \texttt{temperature}: 0. This implies that when the model generates tokens in response to the prompt, it picks the most likely token, given the posterior distribution of candidate tokens. This ensures that the output is as deterministic as possible (there can be some randomness if there are ties).
    \item \texttt{top\_p}: 1. This parameter is essentially irrelevant when temperature is set to 0.
    \item \texttt{n}: 1. This parameter sets the number of responses returned by the OpenAI API in response to the prompt. (Other models do not have this parameter).
\end{itemize}
In all cases, \texttt{messages} is defined as follows: 
\begin{displayquote} 
{\footnotesize
\begin{verbatim}
messages=[{
    "role": "user", 
    "content": f"""{system_message["pre_text"]}\n{text}\n{system_message["post_text"]}"""}
			]
\end{verbatim}
}
\end{displayquote}
where \texttt{text} is the text document to be scaled (e.g., tweet, set of tweets, part of a party manifesto or speech) and \texttt{system\_message} is a Python dictionary with two entries:
\begin{itemize}
    \item   \texttt{pre\_text}: Text to be included in the prompt before the text to be scaled.
    \item \texttt{post\_text}: Text to be included in the prompt after the text to be scaled.
\end{itemize}
Therefore, \texttt{content}  is a string variable that contains the text to be scaled and the scaling instructions submitted to the LLM. In some cases, it is further processed by a utility function to add some special tokens necessary for a particular LLM.

\subsection{OpenAI API}

\begin{itemize}
    \item GPT-4 (\texttt{gpt-4-0613}): We used the following command:
\begin{displayquote} 
{\footnotesize
\begin{verbatim}
response_raw = client.chat.completions.create(
					  model=llm,
					  messages = messages,
					  max_tokens= max_tokens,
					  temperature=temperature, 
    				  top_p = top_p,
					  n = n
					)
\end{verbatim}}
\end{displayquote}

\item GPT-4o (\texttt{gpt-4o-2024-05-13}), 
GPT-4 Turbo (\texttt{gpt-4-turbo-2024-04-09}), GPT-4o mini 
(\texttt{gpt-4o-mini-2024-07-18}) and GPT 3.5 Turbo (\texttt{gpt-3.5-turbo-0125}): We used the following command (a JSON mode is available for these models but not for the earlier models):
\begin{displayquote} 
{\footnotesize
\begin{verbatim}
response_raw = client.chat.completions.create(
    model=llm,
    response_format={ "type": "json_object" },
    messages=messages,
    max_tokens= max_tokens,
    temperature=temperature, 
    top_p = top_p,
    n = n
    )
\end{verbatim}}
\end{displayquote}
\end{itemize}

\subsection{Mistral AI API}

The variable \texttt{messages} is processed by a useful function part of the \texttt{mistralai} Python package before submission to the LLM via the API.
\begin{displayquote} 
{\footnotesize
\begin{verbatim}
messages_Mistral_API = [ChatMessage(role="user", content=messages[0][`content'])]
\end{verbatim}}

\end{displayquote}

\begin{itemize}
    \item Models published in 2023: \texttt{open-mixtral-8x7b}: We used the following command from the \texttt{mistralai} Python package:
\begin{displayquote} 
{\footnotesize
\begin{verbatim}
response_raw = client.chat(
    model=llm,
    messages=messages_Mistral_API,
    max_tokens = max_tokens,
    temperature = temperature,
    top_p = top_p,
    )
\end{verbatim}}
\end{displayquote}

\item Models published in 2024: \texttt{mistral-large-2402}, \texttt{mistral-large-2407}, \texttt{open-mixtral-8x22b}, \texttt{open-mistral-nemo-2407}: We used the following command from the \texttt{mistralai} Python package (a JSON mode is available for those model but not for the models published in 2023):

\begin{displayquote} 
{\footnotesize
\begin{verbatim}
response_raw = client.chat(
    model=llm,
    response_format={ "type": "json_object" },
    messages = messages_Mistral_API,
    max_tokens = max_tokens, 
    temperature=temperature,
    top_p = top_p,
    )
\end{verbatim}}
\end{displayquote}

\end{itemize}

\subsection{Running models with the llama-cpp-python Python library}
Some LLMs are available for download in .gguf format, in particular quantized models (with compressed weights) that can be run locally on the user's computer (with Apple Silicon or a powerful consumer-grade GPU such as Nvidia RTX3090). These models are executed with the llama-cpp-python Python library. Prompts need to be formatted for each model (by adding some special tokens before and after the main message, and that differ across models). To construct the prompt from the message, we used Huggingface's transformer's library. The Huggingface website contains a vast catalogue of LLMs, with model architecture, weights, and associated utilities. By pointing the transformer library to the URL of a particular LLM (\texttt{hf\_path} below), the user can access utility functions adapted to that LLM ({\sf tokenizer.apply\_chat\_template}). We used the following commands:
\begin{displayquote} 
{\footnotesize
\begin{verbatim}
tokenizer = AutoTokenizer.from_pretrained(hf_path)
prompt = tokenizer.apply_chat_template(messages, 
    tokenize=False, 
    add_generation_prompt=True
    )
model_local = Llama(model_path = model_path,
    n_ctx = config_LLM.max_tokens_text_scaling,
    n_gpu_layers = -1,
    use_mlock = True,
    logits_all = False,
    )
response_raw = model_local(
    prompt = prompt, 
    max_tokens = max_tokens, 
    temperature=temperature, 
    top_p=top_p, 
    )
\end{verbatim}
}
\end{displayquote}
Here \texttt{Llama} is a function that loads a model stored locally (at path \texttt{model\_path} into the function \texttt{model\_local}. Then the prompt is submitted to the LLM via this function.

\subsection{Running models on an Apple computer with the mlx-lm Python library}
The \texttt{mlx-lm} Python library is a deep learning framework optimized for Apple computers with `Apple Silicon' processors (e.g. variants of M1, M2, M3). It plays the same role for these processors as that of the widely used Pytorch library for the NVidia GPUs.

The LLMs executed with the \texttt{mlx-lm} Python library were downloaded from HuggingFace.co. With each model comes a tokenizer that, among other things, can convert a prompt to the format adapted to the particular model used for scaling. The prompt query was then submitted to the LLM via the \texttt{generate} function of the \texttt{mlx-lm} library. 
\begin{displayquote} 
{\footnotesize
\begin{verbatim}
model_mlx, tokenizer = load(mlx_path)
prompt = tokenizer.apply_chat_template(
    messages, tokenize=False, add_generation_prompt=True
    )
response_raw = generate(model_mlx, 
    tokenizer, 
    prompt=prompt, 
    max_tokens = max_tokens, 
    temp = temperature, 
    top_p=top_p, 
    verbose=True)	
\end{verbatim}
}
\end{displayquote}
Here the model is stored locally in folder \texttt{mlx\_path}. The function \texttt{load} from the \texttt{mlx-lm} library loads the LLM in the \texttt{model\_mlx} variable and the tokenizer in the \texttt{tokenizer} variable. Then the tokenizer is used to pre-process the messages to create the prompt. And finally, the prompt is submitted to the LLM via the \texttt{generate} function from the \texttt{mlx-lm} library.

\subsection{Running models on an Huggingface inference endpoint}
The Huggingface model sharing platform provides a service called `Inference endpoints' that allows users to run LLMs on dedicated cloud machines provided by Amazon Web Services (AWS), Microsoft Azure or Google Cloud Platform, a location for the machine (e.g., in the US or in Europe). It offers a graphical interface that allows the user to choose an LLM (e.g., Llama 3 70B Instruct), a GPU infrastructure on which to run it (with different amounts of video memory), and various configuration parameters (such as the maximum number of tokens). Once the setup is completed, the user is provided with a custom URL that is then used to send the prompts to the LLM (stored in the variable \texttt{API\_URL} in the code snippet below). The cost is per hour (between 1 USD for a machine with 24GB of video memory that can run models such as Llama 3 8B to 16 USD for a machine with 320GB of video memory that can run models such as Llama 3 70B).

\begin{displayquote} 
{\footnotesize
\begin{verbatim}
headers = {
		"Accept" : "application/json",
		"Authorization": f"Bearer {HF_TOKEN}",
		"Content-Type": "application/json" 
	}

def hf_inference_endpoint_generate(payload):
	response = requests.post(API_URL, headers=headers, json=payload)
		return response.json()

prompt = tokenizer.apply_chat_template(
    messages, tokenize=False, add_generation_prompt=True
    )
    
response_raw = hf_inference_endpoint_generate({
					"inputs": prompt,
					"parameters": {
						"top_p": 0.999,
						"temperature": 0.001,
						"max_new_tokens": max_tokens,
						"do_sample": do_sample,
						"return_text": False,
						"return_full_text": False,
						"return_tensors": False,
						"clean_up_tokenization_spaces": True
					}
				})
\end{verbatim}
}
\end{displayquote}

Here, \texttt{HF\_TOKEN} is a unique identification token users of Huggingface can create in their account settings. The funcion \texttt{hf\_inference\_endpoint\_generate} sends the prompt \texttt{prompt} to the dedicated cloud machine. 

\newpage
\section{Messages submitted to LLMs} 
\label{sec_message_submitted_to_LLMS}%

In all cases, \texttt{messages} is defined as follows: 
\begin{displayquote} 
{\footnotesize
\begin{verbatim}
messages=[{
    "role": "user", 
    "content": f"""{system_message["pre_text"]}\n{text}\n{system_message["post_text"]}"""}
			]
\end{verbatim}}
\end{displayquote}

Next, we report on the values of \texttt{
pre\_text} and \texttt{post\_text} in the various settings. 

\subsection{Tweets}

\subsubsection{pre\_text}
\begin{displayquote}
    You will be provided with the text of a tweet published by a member of the US Congress. Where does this text stand on the `left' to `right' wing scale?  Provide your response as a score between 0 and 100 where 0 means `Extremely left' and 100 means `Extremely right'. If the text does not have political content, set {score} to `NA'. You will only respond with a JSON object with the key Score. Do not provide explanations.
\end{displayquote}

\subsubsection{post\_text}
Left empty.

\subsection{Tweets published by each senator}
\subsubsection{pre\_text}
\begin{displayquote}
    You will be provided with the text of a tweet published by a member of the US Congress. Where does this text stand on the `left' to `right' wing scale?  Provide your response as a score between 0 and 100 where 0 means `Extremely left' and 100 means `Extremely right'. If the text does not have political content, set the score to ``NA''. You will only respond with a JSON object with the key Score. Do not provide explanations.
\end{displayquote}

\subsubsection{post\_text}
Left empty.

\subsection{Sets of tweets published by each senators}
\subsubsection{pre\_text}
\begin{displayquote}
    You will be provided with the text of a set of tweets published by a member of the US Congress. Tweets are separated by a line break followed by the string of characters <TWEET>.
\end{displayquote}

\subsubsection{post\_text}
\begin{displayquote}
    Where does the author of these tweets stand on the `left' to `right' wing scale?  Provide your response as a score between 0 and 100 where 0 means `Extremely left' and 100 means `Extremely right'. If the text does not have political content, set the score to "NA". You will only respond with a JSON object with the key Score. Do not provide explanations.
\end{displayquote}

\subsection{British party manifestos - scaling without explanations about the policy dimensions.}

\subsubsection{pre\_text}
\begin{displayquote}
You will be provided with a sentence from a party manifesto.
\end{displayquote}

\subsubsection{post\_text}
\begin{itemize}
\item Economic policy:
\begin{displayquote}
    Where does this sentence stand on the `left' to `right' wing scale, in terms of economic policy? Provide your response as a score between 0 and 100 where 0 means `Extremely left' and 100 means `Extremely right'. If the sentence does not refer to economic policy, return "NA". You will only respond with a JSON object with the key Score. Do not provide explanations.
\end{displayquote}

\item Social Policy:
\begin{displayquote}
    Where does this sentence stand on the `liberal' to `conservative' scale, in terms of social policy? Provide your response as a score between 0 and 100 where 0 means `Extremely liberal' and 100 means `Extremely conservative'. If the sentence does not refer to social policy, return "NA". You will only respond with a JSON object with the key Score. Do not provide explanations.
\end{displayquote}

\end{itemize}

\subsection{British party manifestos - scaling with explanations about the policy dimensions.}
\label{app: British party manifestos - scaling with explanations about the policy dimension}
\subsubsection{pre\_text}
\begin{displayquote}
You will be provided with a sentence from a party manifesto.
\end{displayquote}

\subsubsection{post\_text}

We used \texttt{post\_text} that included explanations about the scale dimension and the two ends of the scale designed to be as close as possible to the instructions given to crowdworkers in \cite{benoit2016crowd}.

\begin{itemize}

\item Economic policy:
\begin{displayquote}
    Where does this sentence stand on the `left' to `right' wing scale, in terms of economic policy? 
    
    ``Economic'' policies deal with all aspects of the economy, including:
    
    Taxation;
    
    Government spending;
    
    Services provided by the government or other public bodies;
    
    Pensions, unemployment and welfare benefits, and other state benefits;
    
    Property, investment and share ownership, public or private;
    
    Interest rates and exchange rates;
    
    Regulation of economic activity, public or private;
    
    Relations between employers, workers and trade unions.
    
    ``Left'' economic policies tend to favor one or more of the following:
    
    High levels of services provided by the government and state benefits, even if this implies high levels of taxation;
    
    Public investment. Public ownership or control of sections of business and industry;
    
    Public regulation of private business and economic activity;
    
    Support for workers/trade unions relative to employers.

    ``Right'' economic policies tend to favor one or more of the following:
    
    Low levels of taxation, even if this implies low levels of levels of services provided by the government and state benefits;
    
    Private investment. Minimal public ownership or control of business and industry;
    
    Minimal public regulation of private business and economic activity;
    
    Support for employers relative to trade unions/workers.
    
    Provide your response as a score between 0 and 100 where 0 means `Extremely left' and 100 means `Extremely right'. If the sentence does not refer to economic policy, return "NA". You will only respond with a JSON object with the key Score. Do not provide explanations.
\end{displayquote}

\item Social Policy:
\begin{displayquote}
    Where does this sentence stand on the `liberal' to `conservative' scale, in terms of social policy? 
    
    ``Social'' policies deal with aspects of social and moral life, relationships between social groups, and matters of national and social identity, including:
    
    Policing, crime, punishment and rehabilitation of offenders;
    
    Immigration, relations between social groups, discrimination and multiculturalism;
    
    The role of the state in regulating the social and moral behavior of individuals.

    ``Liberal'' social policies tend to favor one or more of the following:
    
    Recording the contents of political text on economic and social scales;
    
    Policies emphasizing prevention of crime, rehabilitation of convicted criminals;
    
    The right of individuals to make personal moral choices on matters such as abortion, gay rights, and euthanasia;
    
    Policies penalizing discrimination against particular social groups and/or favoring a multicultural society.

    ``Conservative'' social policies tend to favor one or more of the following:
    
    Policies emphasizing more aggressive policing, increasing police numbers, conviction and punishment of criminals, building more prisons;
    
    The right of society to regulate personal moral choices on matters such as abortion, gay rights, and euthanasia;
    
    Policies favoring restriction of immigration, and/or opposing explicit provision of state services for minority cultures.
    
    Provide your response as a score between 0 and 100 where 0 means `Extremely liberal' and 100 means `Extremely conservative'. If the sentence does not refer to social policy, return "NA". You will only respond with a JSON object with the key Score. Do not provide explanations.
\end{displayquote}

\end{itemize}

\subsection{EU legislative speeches - scaling with background explanations.}

We used the following \texttt{pre\_text}  that included background information about the context of the legislative debate (especially the nature of the proposal being discussed), the scale dimension and the two ends of the scale designed to be as close as possible to the instructions given to crowdworkers in \cite{benoit2016crowd}:

\subsubsection{pre\_text}
\begin{displayquote}
\#\# Summary
This task involves reading a sentence from a debate over policy in the European parliament, and judging whether particular statements were for or against a proposed policy.

\#\# Background.

The debates are taken from a debate in the European Parliament over the ending of state support for uncompetitive coal mines.
In general, state aid for national industry in the European Union is not allowed, but exceptions are made for some sectors such as agriculture and energy. At stake here were not only important policy issues as to whether state intervention is preferable to the free market, but also the specific issue for some regions (e.g. Ruhrgebiet in Germany, the north-west of Spain, the Jiu Valley in Romania) where the social and economic impacts of closure would be significant, possibly putting up 100,000 jobs at risk when related industries are considered.
	
Specifically, the debate concerned a proposal by the European Commission to phase out all state support by 2014. Legislation passed in 2002 that allowed for state subsidies to keep non- profitable coal mines running was due to end in 2010. The Commission proposed to let the subsidies end, but to allow limited state support until 2014 in order to soften the effects of the phase-out. A counter proposal was introduced to extend this support until 2018, although many speakers took the opportunity to express very general positions on the issue of state subsidies and energy policy.

\#\# Your Coding Job.

Your key task is to judge individual sentences from the debate according to which of two contrasting positions they supported:
\begin{itemize}
    \item[-] Supporting the rapid phase-out of subsidies for uncompetitive coal mines. This was the essence of the council proposal, which would have let subsides end while offering limited state aid until 2014 only.
    \item[-] Supporting the continuation of subsidies for uncompetitive coal mines. In the strong form, this involved rejecting the Commission proposal and favoring continuing subsidies indefinitely. In a weaker form, this involved supporting the compromise to extend limited state support until 2018.
\end{itemize}

\#\# The `pro-subsidy’ to ‘anti-subsidy’ scale.

The ``anti-subsidy" position was the essence of the Commission proposal, which would have let subsidies end while offering limited state aid until 2014 only. 
Examples of ``anti-subsidy" positions: 

	Statements declaring support for the commission position.
 
	Statements against state aid generally, for reasons that they distort the market.
 
	Arguments in favor of the Commission phase-out date of 2014, rather than 2018.

In the strong form, the ``pro-subsidy" position, involved rejecting the Commission proposal and favoring continuing subsidies indefinitely. In a weaker form, this involved supporting the compromise to extend limited state support until 2018. 

Examples of ``pro-subsidy" positions: 
	 Arguments that keeping the coal mines open to provides energy security.
  
	 Arguments that coal mines should be kept open to provide employment and other local economic benefits.
  
	 Preferences for European coal over imported coal, for environmental or safety reasons.
  
\#\# Sentence:

\end{displayquote}

\subsubsection{post\_text}

\begin{displayquote}
\#\#  Instructions.

Where does this sentence stand on the `anti-subsidy’ to ‘pro-subsidy’ scale? Provide your response as a score between 0 and 100 where 0 means `Extremely anti-subsidy' and 100 means `Extremely pro-subsidy.' If the sentence does not refer to subsidy policy, return "NA". You will only respond with a JSON object with the key Score. Do not provide explanations.
\end{displayquote}

\newpage
\section{Other Scaling Methods} 
\subsection{British party manifestos - fine-tuned BERT classifier}
Note: We used a Python Jupyter notebook executed on Google Colab adapted from that made available in the companion OSF folder to the paper on using machine learning to measure typicality by \cite{LeMens2023SociologicalScience_BERT_typicality} (\href{https://osf.io/ta273/}{https://osf.io/ta273/}).

We implemented a BERT classifier that would predict the sentence coding crowdworkers in \cite{benoit2016crowd}. For the manifestos used for scaling on the Economic and Social policy dimensions, we specified an 11 class sentence classifier in which the classes were the 5 positions on the Economic policy dimension [-2,-2,0,1,2] the 5 positions on the Social policy dimension [-2,-2,0,1,2], and `nan' in case a crowdworker judged a sentence to be neither about economic nor social policy. We divided the data into training, validation and prediction sets, ensuring that all ratings for a given sentence were in just one of these three sets. We used 45\% of the sentences for training, 5\% for validation, and 50\% for prediction.  This produced a training set with 98K human ratings, a validation set with 10K human ratings, and a test set with 107K human ratings.

We implemented our BERT model using the \texttt{bert-base-uncased}, the TensorFlow machine-learning library and its higher-level wrapper Keras. We used the Adam optimizer to minimize the cross-entropy loss function to fine-tuning the model. This amounts to trying to find the parameters that maximize the likelihood of the true categories in the training data. We used Google Colab with a TPU-equipped virtual machine to train the model. 

 We used the following (standard) parameter values:
\begin{itemize}
    \item Batch size: 256
    \item Max number of tokens: 512 (this is the maximum possible with \texttt{bert-base-uncased})
    \item Optimizer: adam
    \item Loss function: categorical\_crossentropy
    \item Learning rate: 2e-5
    \item Epochs: 200 (we used a training routine that would treat this parameter as the maximum possible number of epochs - it automatically stopped the training when the validation loss started to increase)
\end{itemize}

We applied the fine-tuned model on the sentences in the prediction set. This gave a vector of categorization probabilities in the 11 candidate classes. To obtain the policy area of the focal sentence, we first added the categorization probabilities for the 5 possible dimensions of the economic policy dimension, and did the same for the Social Policy dimension. This gives a vector of 3 probabilities for Economic Policy, Social Policy and `Other'. We then assigned the focal sentence to the most likely policy area (including `Other'). 

To obtain the position on the economic policy dimension, of a sentence about Economic policy (as per the result of the classification in policy areas described above), we took the weighted average of the 5 possible positions on that dimension, where the weights were the categorization probabilities in these 5 classes (rescaled to sum to 1). From this, we constructed a position estimate for each manifesto as the average position of its sentences. 

We followed the same procedure to obtain position estimates of sentences and then manifestos on the Social policy dimension.

\subsection{Tweets - fine-tuned BERT classifier}
Note that for tweets, we did not train the models on data produced by human coders, but instead trained the classifiers to predict the party of the Congress member who published the tweet (Democratic Party or Republican Party), following the approach proposed by \cite{LeMens2023SociologicalScience_BERT_typicality} and implemented on these Twitter data in \cite{LeMens2023_PNASGPT4_Typicality}. We used the approach used in this latter paper by slightly adapting the jupyter python notebooks made available at on the companion OSF folder (\href{https://osf.io/ta273/}{https://osf.io/ta273/}). We used the data from that paper. The training set consisted in close to 1M tweets published by US Congress members during the 116th and 117th Congress sessions.

For the tweets published by the senators of the 117th US congress, we used as a training set the tweets published by senators of the 116th and 117th Congress sessions, ensuring that none of the tweets in the prediction set (100 tweets per senator) was in the training or validation set. This resulted in a training set of about 246K tweets. We used the categorization probabilities in the two parties to compute the typicality of each tweet in each party (following \cite{LeMens2023SociologicalScience_BERT_typicality}) and took the difference in typicalities in the Republican and Democratic parties as the position of the tweet on the left-right ideological spectrum. The position estimate of a senator is the average position of the 100 tweets in the prediction set. 

For tweets published after the training cut-off date of GPT-4, we used the same training set as \cite{LeMens2023_PNASGPT4_Typicality} and simply computed the position of each tweet following the approach explained in the previous paragraph.

 We used the following (standard) parameter values:
\begin{itemize}
    \item Batch size: 256
    \item Max number of tokens: 512 (this is the maximum possible with \texttt{bert-base-uncased})
    \item Optimizer: adam
    \item Loss function: categorical\_crossentropy
    \item Learning rate: 2e-5
    \item Epochs: 200 (we used a training routine that would treat this parameter as the maximum possible number of epochs - it automatically stopped the training when the validation loss started to increase)
\end{itemize}

\subsection{Tweets - fine-tuned GloVe classifier}
Note: We used a Python Jupyter notebook executed on Google Colab adapted from that made available in the companion OSF folder to the paper on using machine learning to measure typicality by \cite{LeMens2023SociologicalScience_BERT_typicality} (\href{https://osf.io/ta273/}{https://osf.io/ta273/}).
The following text is a minor adaptation of the description of the fine-tuned GloVe classifier in that paper. 

We adapted the approach described in \cite{LeMens2023SociologicalScience_BERT_typicality} to the present empirical setting. We used the same model structure as for the BERT classifier, but we used GloVe word embeddings as a language representation instead of the BERT language representation. More precisely, we used a word embedding layer with pre-trained weights in our classifiers instead of the BERT language representation.  We used GLOVE word embeddings \citep{pennington2014glove} to transform text documents into vectors. GLOVE is a \emph{word}-embedding model, not a text-embedding model. Consequently, we needed to combine word positions in the embedding space to create a unique position for text documents. We used the average position of the words in the text as the position of the tweet in semantic space. 

We selected the top 20,000 most frequent words in the training data, then we used the Glove embedding model ``glove.6B.300d'' to transform each of the 20,000 words into vector positions. With this embedding, we created a basic deep-learning model consisting of:
\begin{enumerate}
    \item Embedding layer
    \item Pooling 1D layer
    \item Dense layer (with softmax activation)
\end{enumerate}
We trained the model for up to 1,000 epochs with a $2e-3$ learning rate using a categorical cross entropy loss function and a batch size of 16,384.

\subsection{Tweets - fine-tuned Naive Bayes classifier with tf-idf language representation}
Note: We used a Python Jupyter notebook executed on Google Colab adapted from that made available in the companion OSF folder to the paper on using machine learning to measure typicality by \cite{LeMens2023SociologicalScience_BERT_typicality} (\href{https://osf.io/ta273/}{https://osf.io/ta273/}).
The following text is a minor adaptation of the description of the naive Bayes classifier in that paper. 

We also used a standard machine-learning text classifier based on a version of the Bag-of-Words representation that weighs word frequencies by diminishing the importance of words that occur in many text documents. This approach is known as `TF-IDF', or `term frequency-inverse document frequency', \citep{jones1972statistical}. The classifier model is the Naive Bayes classifier \citep{maron1961automatic}. This machine-learning classifier produces categorization probabilities based on word co-occurrences. It is computationally undemanding, but its representation of text documents is not sensitive to the order of words in sentences. Also, the representation of words does not depend on their semantic similarity. 

To train the classifier applied on the tweets of the senators of the 117th US Congress (Section~\ref{sub_results_congress_117_senators}), we selected the 5,000 more frequent words from the training data (this number is constrained by the RAM of the computer on which we ran the model training, which in this case was a virtual machine on Google Colab with 50GB of RAM). We assigned an ID to each word and transformed all the book descriptions using this ID dictionary. Finally, we fit a multinomial Naive Bayes on all the book descriptions in the training data. 

To train the classifier applied to the tweets published after the GPT-4 training cut-off date (Section~\ref{sub_results_congress_118_crowdsourcing}), we selected the 2,000 more frequent words from the training data (this number is lower than in the previous paragraph because the training data is larger in this case and thus running the model training requires more RAM for the same vocabulary size). 

For the construction of the dictionary, we used the sklearn package ``CountVectorizer'' and to fit the model we used sklearn package ``MultinomialNB.''

\newpage
\section{Additional Results - Tweets published by members of the US Congress after the training cut-off date of GPT-4}\label{sec_additional_results_tweets_congress_118}

\FloatBarrier

Visual comparison of the performance of the various scaling methods (LLMs, fine-tuned BERT classifier, fine-tuned GloVe classifier, and naive Bayes classifier with tf-idf representation).

\begin{figure}[htbp]
    \centering
    \includegraphics[width = \columnwidth]{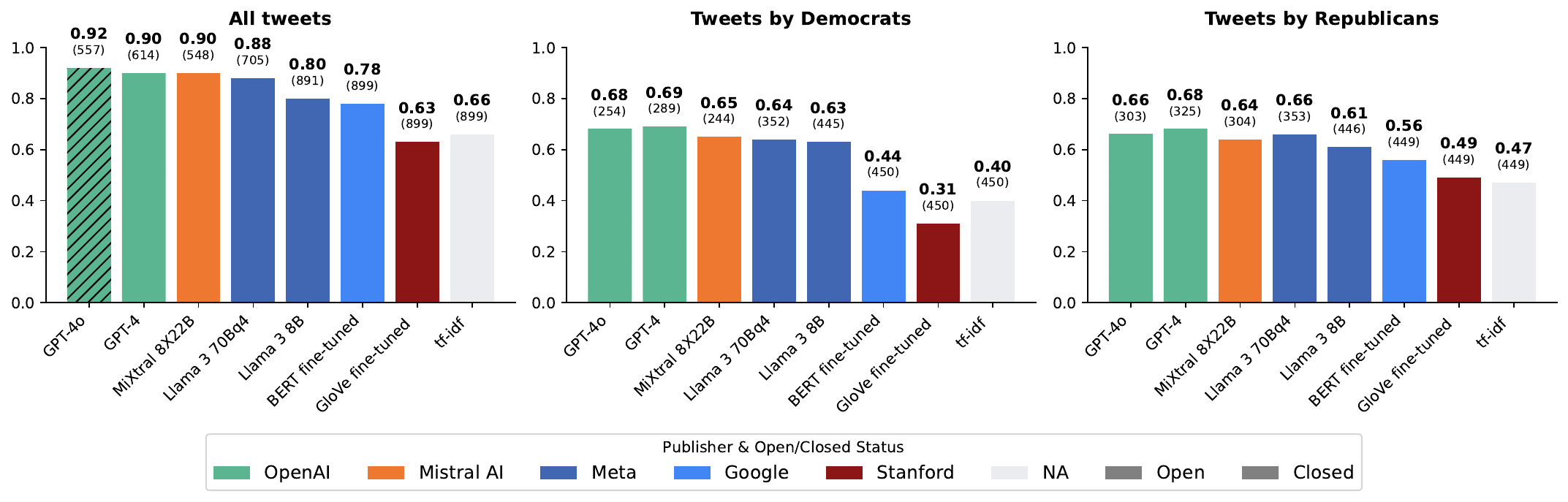}
    \caption{Positioning tweets published by members of the US Congress after the training cut-off date of GPT-4 on the left-right ideological spectrum ($N=899$).  Pearson correlations between the benchmark (crowdsourced position estimates) and position estimates obtained with LLMs. The numbers in parentheses indicate the number of documents for which a position estimate was obtained (position estimates are missing if the LLM did not return a position score for any of the parts of manifestos submitted to it). The data are the same as for Figure~\ref{fig_tweets_after_cutoff_baseline}.
    }    \label{fig_tweets_after_cutoff_baseline_bar_graphs}
    \vspace{0.1cm}
\end{figure}

The LLMs failed to return position scores for some of the tweets, possibly because these did not include enough political content. To verify that this pattern of non-response is reasonable, we computed the average number of human ratings for the two sets of tweets, for each LLM (survey respondents had the option to respond `Not applicable' to the question about the position of each tweet on the Left-Right ideological spectrum, if they felt the tweet did not contain enough information for them to return a position rating). Under the hypothesis that an LLM fails to return a numeric score for tweets that did not contain enough political content, but would do so for tweets than contain enough political content, we expected that the average number of human position ratings would be larger for tweets that received a position score by the LLM than for tweets that did not receive a position score. These quantities are reported in the last two columns of table~\ref{tab_USCongress_118_body_nb_responses_appendix}. For all the LLMs, the pattern is consistent with the hypothesis.

\begin{table}[htbp]
    \centering
    \includegraphics[width = \columnwidth]{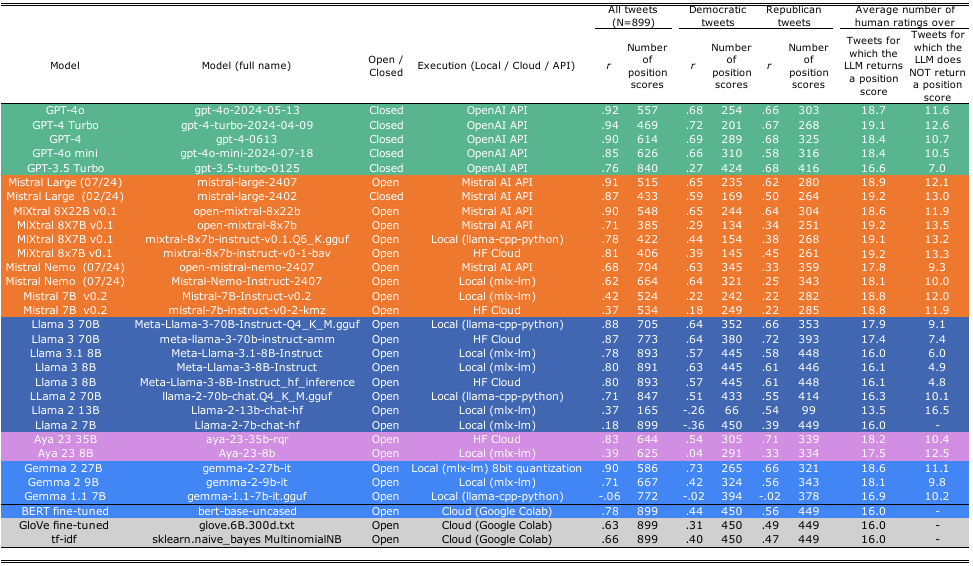}
    \caption{Positioning tweets published by members of the US Congress after the training cut-off date of GPT-4 on the left-right ideological spectrum ($N=899$).  The last two columns report the average number of human position ratings for tweets that received a position score, and the same quantity for those for which the LLM did not return a position score. This table includes the LLMs from the body of the paper as well as a set of additional LLMs (from the previous generation). `Open' LLMs can be downloaded on the user's machine and run locally if the machine has enough video memory (under more or less restrictive licenses such as Apache 2.0, the Meta Llama 3 community license agreement, or the Mistral AI Research License).
   `Local' execution means that the model was run on our laptop after downloading the model `API' execution means that the model was run for a per-token fee on a machine managed by the provider of the model. `Cloud' execution means that the model was run via a cloud computing service (Google Colab or Huggingface Inference Endpoints) on a machine equipped with a TPU or GPUs (payment per hour / compute unit). 
   Models in .gguf format are versions of the model with compressed (`quantized') weights and were run locally with the \texttt{llama-cpp-python} package. In the full names of these models,`QX' indicates the number of bits used to represent the model weights. We used minimal compression that would allow us to run the model on our laptop (an Apple Macbook Pro with an M1 Max processor with 64GB that can run models as large as 40GB). Llama 3.1 8B, Llama 3 8B, Llama 2 7B, Llama 2 13B, Gemma 1.1 7B, Gemma 2 9B, Aya 23 8B were downloaded from Huggingface.co and were run locally without any compression with the \texttt{mlx-lm} Python package. For Gemma 2 27B, we used the \texttt{mlx-lm} Python package to apply an 8-bit quantization convertion and to run it locally.
    }    \label{tab_USCongress_118_body_nb_responses_appendix}
    \vspace{0.1cm}
\end{table}


\subsection{Distribution of position estimates produced by LLMs and other scaling methods}
 \begin{figure}[htbp]
    \centering
    \includegraphics[width = .90\columnwidth]{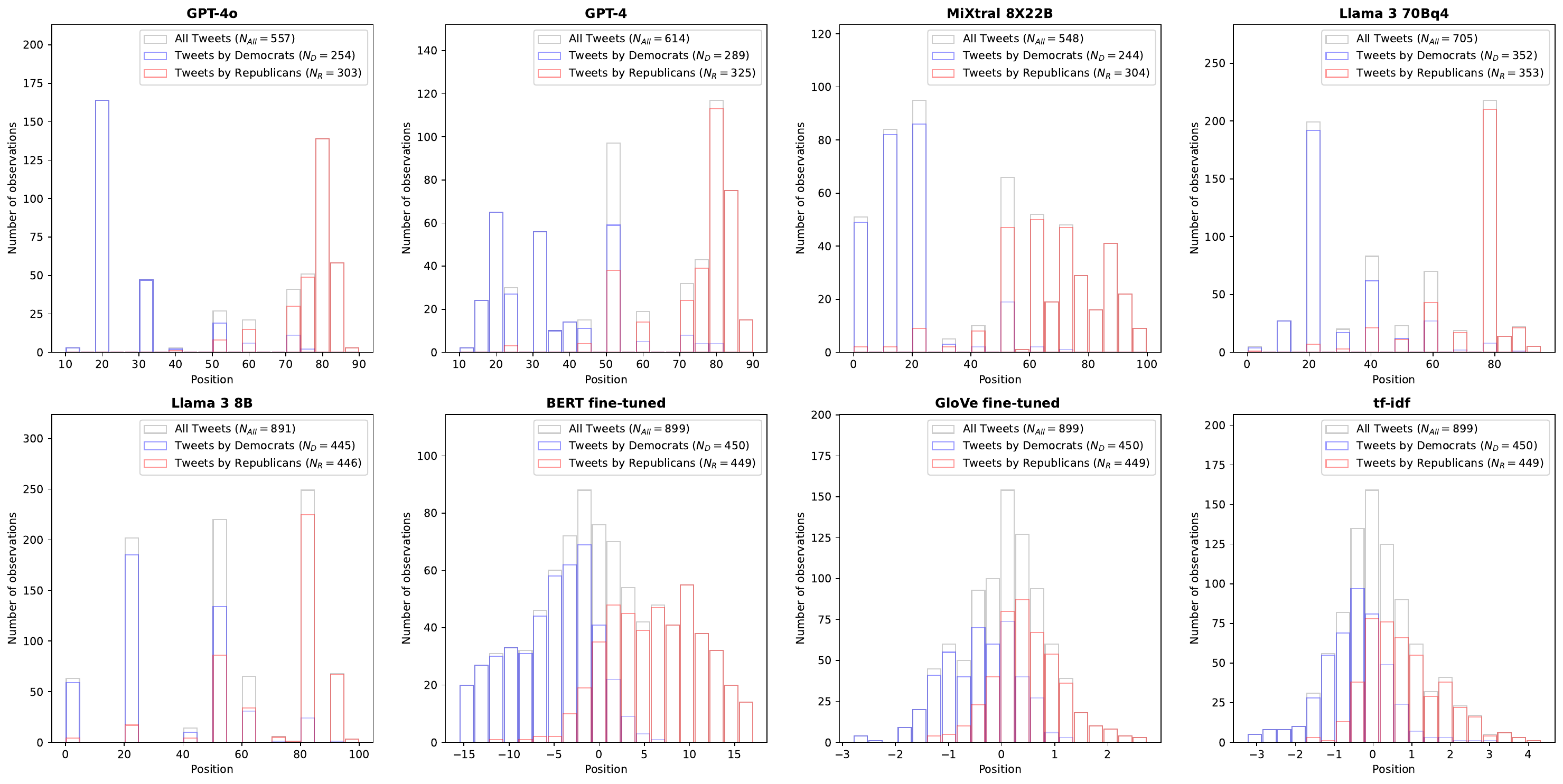}
    \caption{Tweets published by members of the 118th US Congress: Histograms of position estimates produced by the LLMs and other scaling methods.}.
    \label{congress_118_tweets_histograms_LLMs}
    \vspace{0.1cm}
\end{figure}

 \begin{figure}[htbp]
    \centering
    \includegraphics[width = .90\columnwidth]{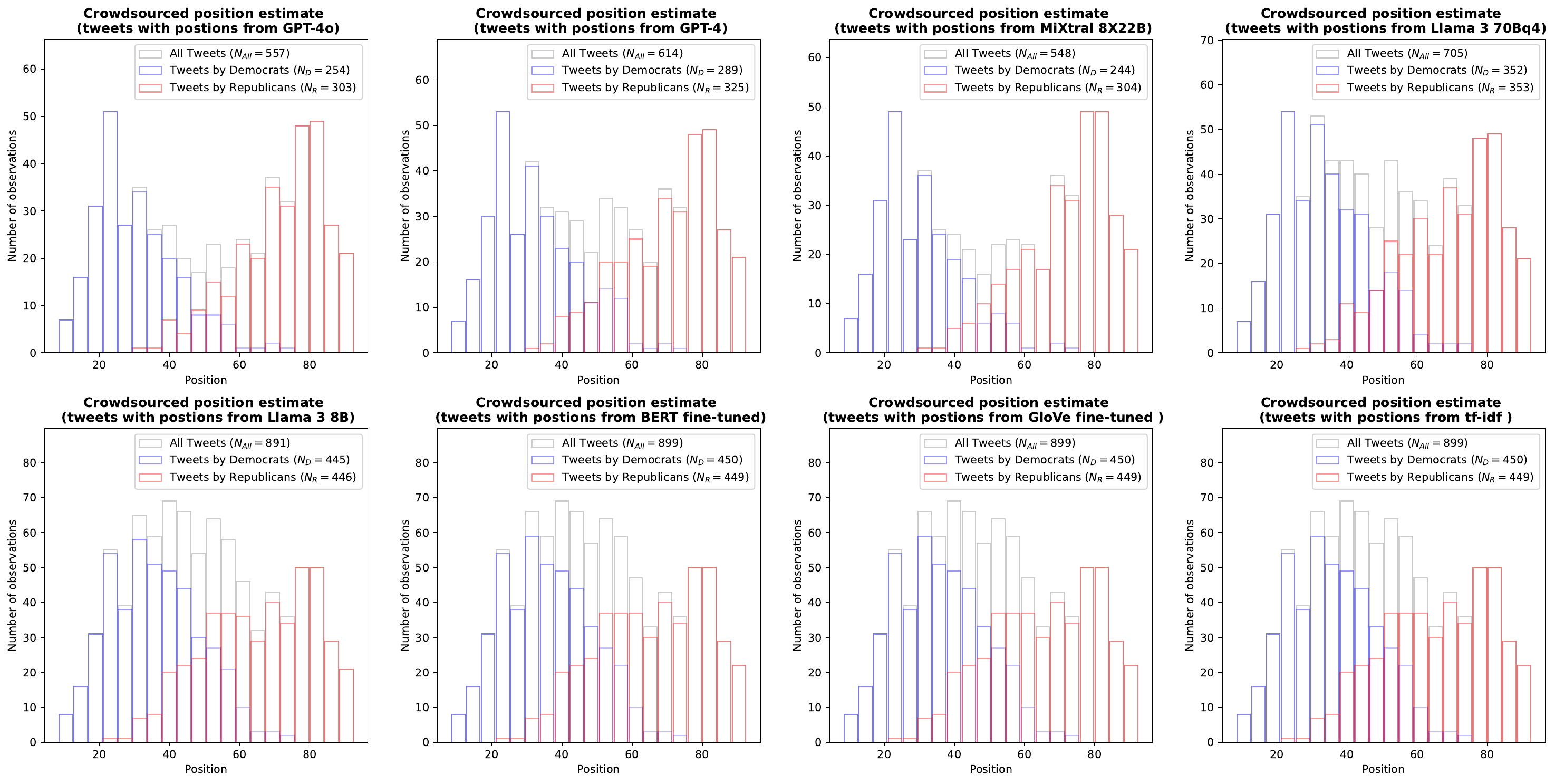}
    \caption{Tweets published by members of the 118th US Congress: Histograms of crowdsourced position estimates for the subset (out of the total of 899 tweets) for which each scaling method produced a position estimate).}
    \label{congress_118_tweets_histograms_crowd_sourced_estimates_subset_tweets_scaled_by_LLMs}
    \vspace{0.1cm}
\end{figure}

\FloatBarrier
\vfill
\pagebreak
\subsection{Scaling tweets on the left-right ideological spectrum by using LLMs to obtain their typicalities in the Democratic and Republican parties}\label{App_delta_typicality_LLMs}

We implemented the approach advocated in \cite{LeMens2023_PNASGPT4_Typicality} to obtain typicality measures of tweets in political parties (and illustrated in that paper with GPT-4). This involves directly asking the LLMs for the typicality of each tweet in each party. We then construct a position estimate of each tweet on the left-right ideological spectrum as the difference between its typicality in the Republican party and its typicality in the Democratic party. \cite{LeMens2023_PNASGPT4_Typicality} had found that the typicality measures obtained with GPT-4 had higher correspondence with human typicality ratings than those produced with other methods, such as a BERT classifier or classifiers based on text embeddings (ADA2 embeddings published by OpenAI) or text embeddings (GloVe) that were trained on vast amounts of data.

\subsubsection{Messages submitted to LLMs.}
We used the same message structure as explained in~\ref{sec_message_submitted_to_LLMS} except for a change to the `\texttt{pre\_text}' part of the user messages.

\subsubsection{pre\_text for typicality in Democratic Party.}
\begin{displayquote}
   You will be provided with the text of a tweet published by a member of the US Congress. How typical is this tweet of the Democratic Party? Provide your response as a score between 0 and 100 where 0 means `Not typical at all' and 100 means `Extremely typical'. You will only respond with a JSON object with the key Score. Do not provide explanations.
\end{displayquote}

\subsubsection{pre\_text for typicality in Republican Party.}
\begin{displayquote}
   You will be provided with the text of a tweet published by a member of the US Congress. How typical is this tweet of the Republican Party? Provide your response as a score between 0 and 100 where 0 means `Not typical at all' and 100 means `Extremely typical'. You will only respond with a JSON object with the key Score. Do not provide explanations.
\end{displayquote}

\subsubsection{post\_text.}
Left empty.

\subsubsection{Results.}
The correlations with crowdsourced position estimates are similar to those obtained with the baseline approach reported in the body of the article, with some gains in terms of within-party correlation. Another key difference is that, with this approach, LLMs return position estimates for all tweets. It should be noted that there is no need for a tweet to have explicitly political content for an artificial or human judge to evaluate its typicality in a political party.  This is because of a fundamental difference between this approach of constructing a position estimate and the baseline approach used in the body of the paper: the baseline approach instructs the LLM to focus on a particular dimension. The approach consists in taking the difference between typicalities in two concepts does not require the specification of a particular dimension: it amounts to a sort of projection on the direction of the vector that connects that centers of the two concepts in multidimensional semantic space. 

\begin{figure}[htbp]
    \centering
    \includegraphics[width = \columnwidth]{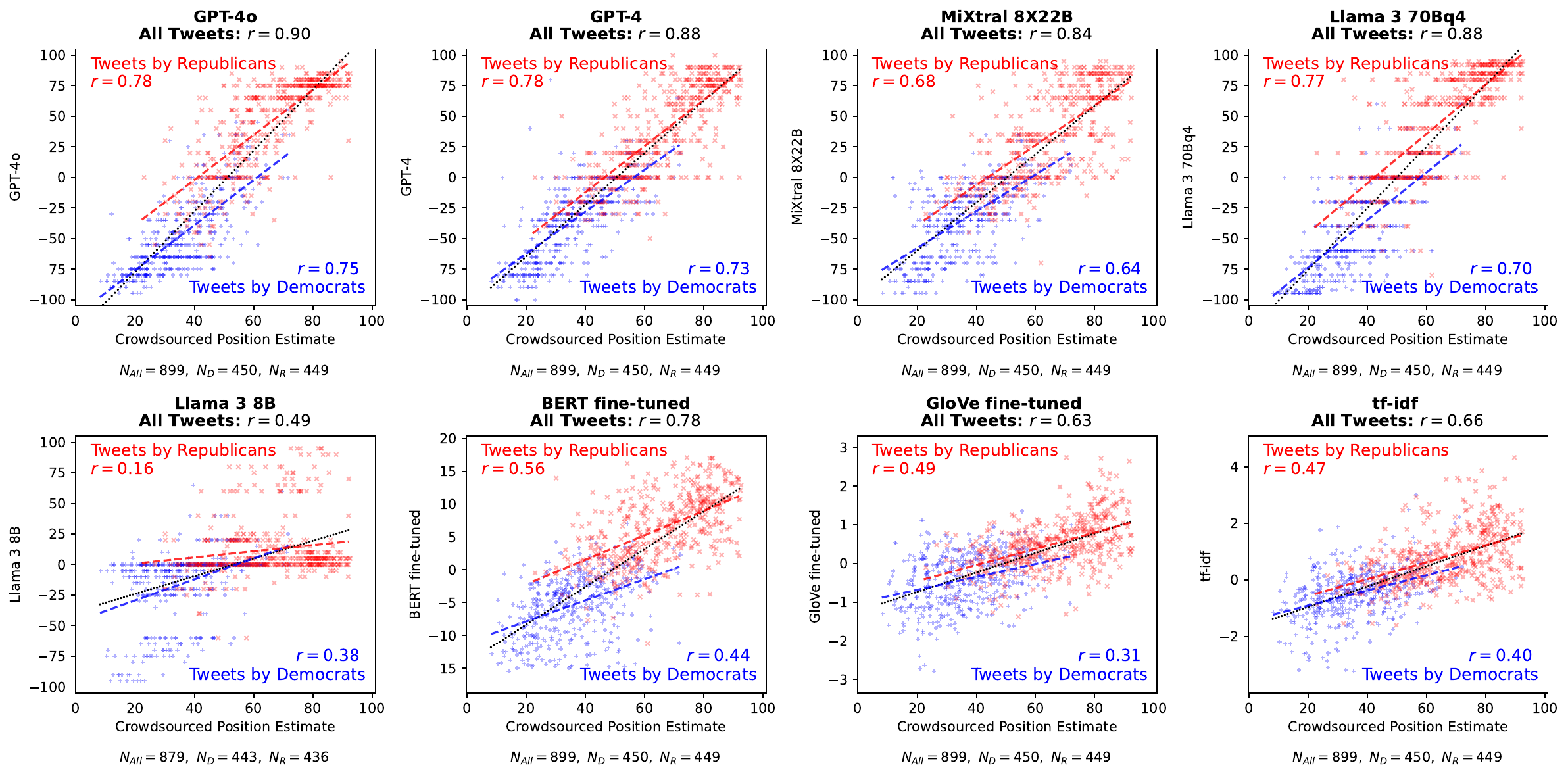}
    \caption{Positioning tweets published by members of the US Congress after the training cut-off date of GPT-4 on the left-right ideological spectrum ($N=899$) by asking for the typicality in the Republican party, their typicality in the Democratic party and defining the position estimate as the difference between these typicalities. The y-axes report the position of each tweet on the left-right ideological spectrum obtained with the various models. The x-axes reports crowdsourced estimates as the average position provided by participants in an online survey of US residents.
      }    \label{fig_tweets_after_cutoff_typ_delta}
    \vspace{0.1cm}
\end{figure}

 \begin{figure}[htbp]
    \centering
    \includegraphics[width = \columnwidth]{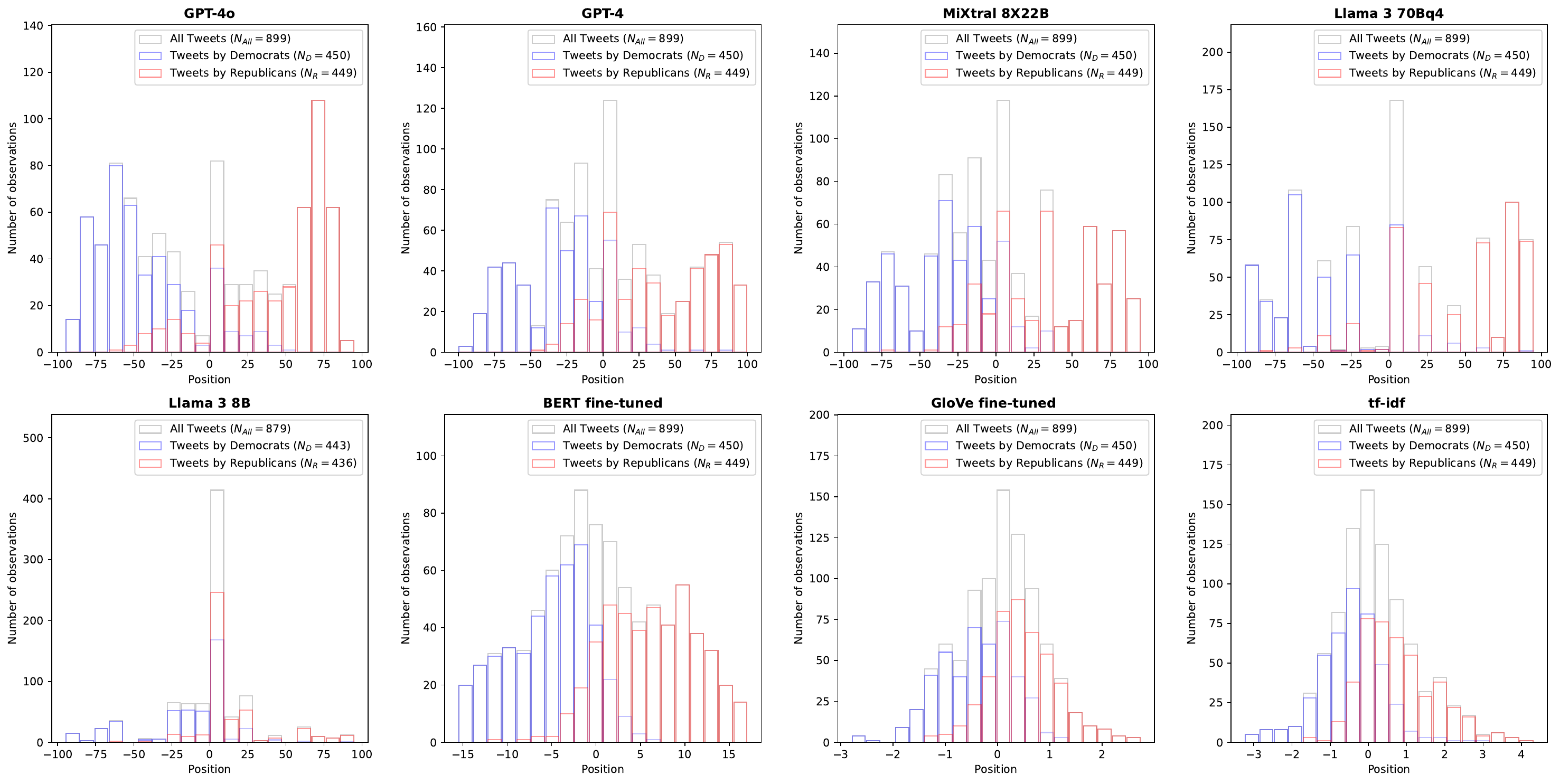}
    \caption{Tweets published by members of the 118th US Congress: Histograms of position estimates produced by the LLMs by taking the difference between the typicality in the Republican party and typicality in the Democratic party.}.
    \label{congress_118_tweets_histograms_LLMs_typ}
    \vspace{0.1cm}
\end{figure}

\FloatBarrier

\subsection{Predicting the position estimates given by one human respondent to the survey of tweet positions: ENO scores} \label{app_sec_ENO_scores}

One existing characterization of the predictive performance of models of human judges focuses on predicting the ratings of one human respondent \citep{erev2007learning}. It asks how many observations of other human participants are necessary to make a prediction as good as the model. The resulting number is the model's equivalent number of observations (ENO). 

Here, we predict the position ratings of one human participant based on the average of $N$ position ratings of other participants. We compute the Pearson correlation between these two quantities. The resulting correlation will depend on the specific ratings selected as criterion and predictor variables. So, for each tweet, we randomly sample one human rating as the criterion variable and treat the remaining ratings as potential predictors. We repeat this procedure 100 times and report the average correlation in Table~\ref{tab_USCongress_118_ENO_tweets}. We do this for tweets that received at least 15 human position ratings (at least 15 respondents in our survey did not select `NA' for that tweet), which leads to a set of 598 tweets. We repeat the procedure for values of $N$ that range from 1 to 14 (the number of human typicality ratings minus 1), and over three sets of tweets: tweets from Democratic and Republican politicians, tweets from Democratic politicians and tweets from Republican politicians.

It should be noted that the column for N=1 reports the pairwise correlation between the position ratings given by two randomly picked human judge, computed over the tweets for which the focal model returned a position score.  Because the set of tweets for which each model returned a position score differs across model, this measure of pairwise correlation differs across rows in the column. The same observation applies to the other values of N. 

The ENO scores for the tweet position estimates provided by the LLMs are all higher than 1 and go up to 5 to 7 (for GPT-4, MiXtral 8X22B and Llama 3 70Bq4, over all tweets). In the latter case, this means that the position scores provided by these models are as good predictors as the average position ratings given by 5 independent human participants in our survey of Prolific participants who reside in the US.

\begin{table*}
    \centering
    \caption{Tweet data: Predicting one human position rating rating with position ratings by other human coders and with the position estimates provided by the LLMs) }
        \includegraphics[width = \textwidth]{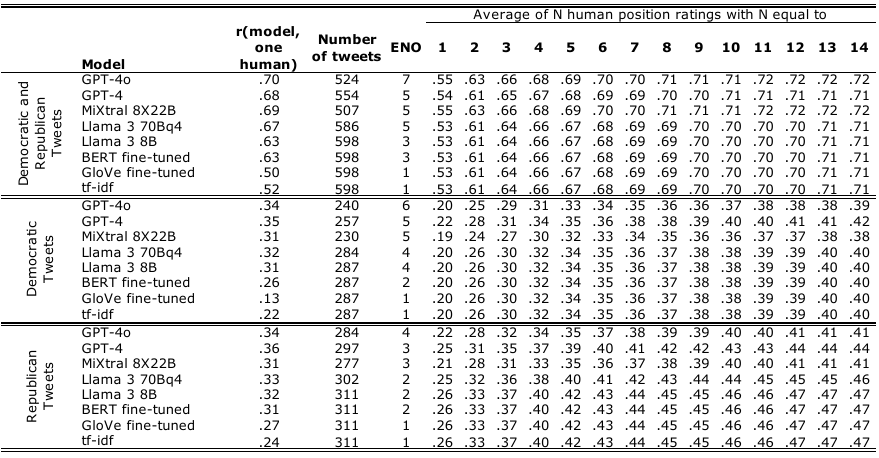}
   \label{tab_USCongress_118_ENO_tweets}
\end{table*}

\FloatBarrier

\newpage
\section{Additional Results - Senators of the 117th US Congress}
Visual comparison of the performance of the various scaling methods (LLMs, fine-tuned BERT classifier, fine-tuned GloVe classifier, and naive Bayes classifier with tf-idf representation).

\FloatBarrier

 \begin{figure}[htbp]
    \centering
    \includegraphics[width = \columnwidth]{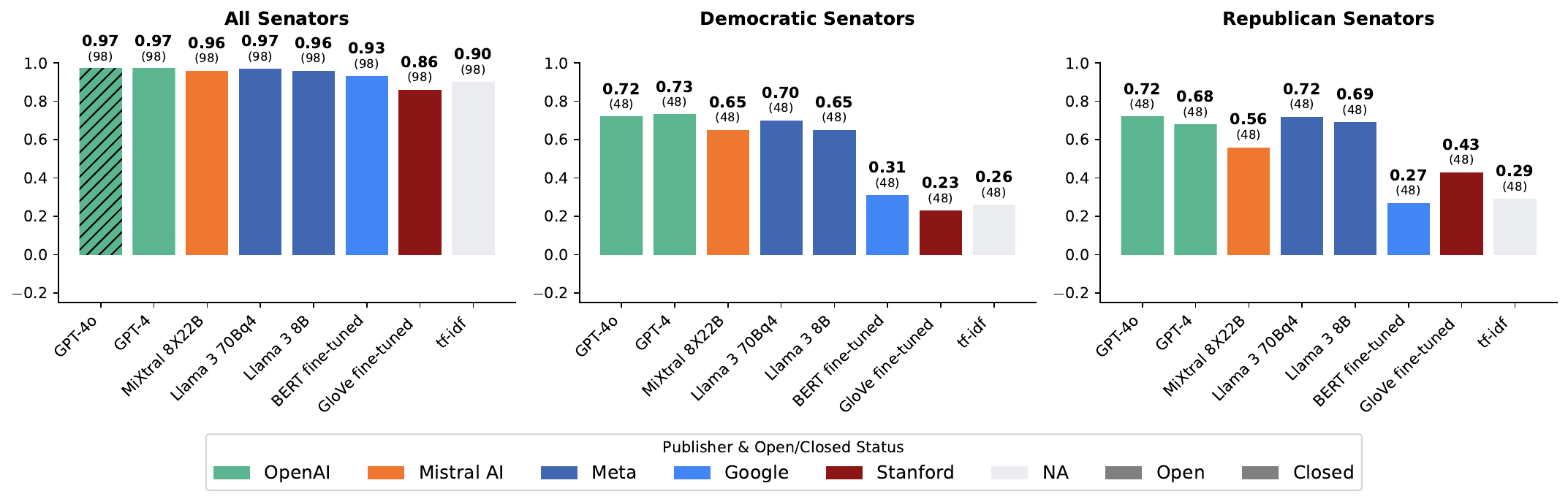}
    \caption{Positioning Senators of the 117th Congress on the left-right ideological spectrum based on their tweets ($N=98$). Pearson correlations between the benchmark (First dimension Nokken-Poole period-specific DW-NOMINATE score) and position estimates obtained with LLMs and other scaling methods. The numbers in parentheses indicate the number of senators for which a position estimate was obtained. The data are the same as for Figure~\ref{fig_senator_117_sentence_scaling_Nokken_Poole}.
    }    \label{fig_senator_117_tweet_scaling_bar_graphs_Nokken_Poole}
    \vspace{0.1cm}
\end{figure}

\FloatBarrier

\subsection{Correlation between the position estimates of the senators (using the baseline approach described in the body of the article) and a position estimate measure based on campaign funding.}
 \begin{figure}[htbp]
    \centering
    \includegraphics[width = \columnwidth]{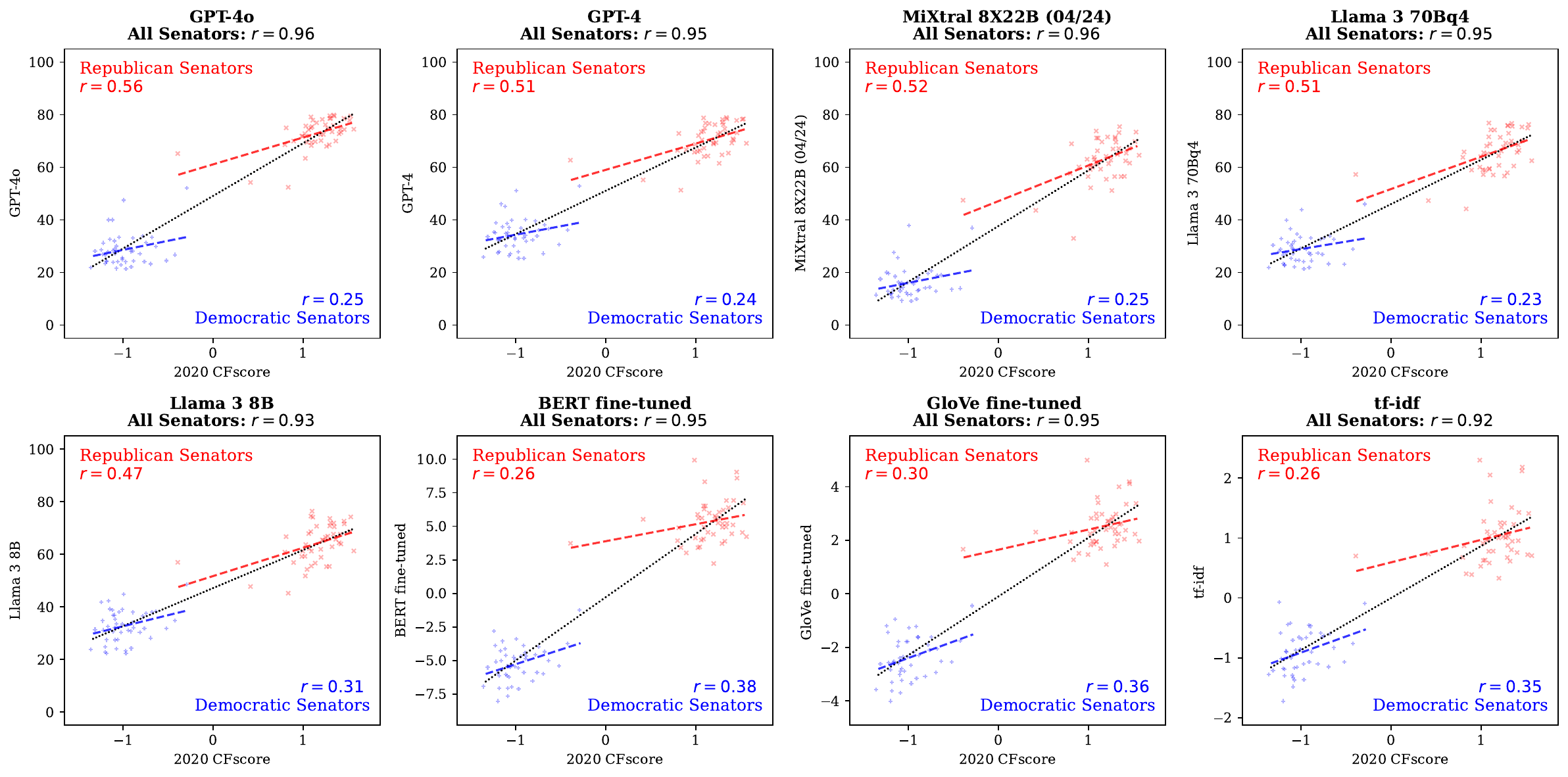}
    \caption{Replication of Figure~\ref{fig_senator_117_sentence_scaling_Nokken_Poole} using the 2020 CFscore from the Dime database \citep{bonica2014mapping} instead of the period-specific DW-NOMINATE score. The 2020 CFscore is a period-specific estimate of ideology based on donation received.
    }    \label{fig_senator_117_sets_of_tweets_CF_score_dyn}
    \vspace{0.1cm}
\end{figure}

\FloatBarrier

\subsection{Scaling the aggregated text of the 100 tweets in one prompt}
\label{app_senators_one_prompt}
We used the user message to obtain the position of each senator based on the text of the 100 tweets:
\begin{displayquote} \small
You will be provided with the text of a set of tweets published by a member of the US Congress. Tweets are separated by a line break followed by the string of characters <TWEET>.

<TWEET> $\ll$ Text of tweet 1 $\gg$

...

<TWEET> $\ll$  Text of tweet N $\gg$

Where does the author of these tweets stand on the `left' to `right' wing scale?  Provide your response as a score between 0 and 100 where 0 means `Extremely left' and 100 means `Extremely right'. If the text does not have political content, set the score to ``NA''. You will only respond with a JSON object with the key Score. Do not provide explanations.
\end{displayquote}

For the models we used, this resulted in a prompt that could fit the maximum context window. This means the prompt could be processed by the model in a single query

\begin{figure}[htbp]
    \centering
    \includegraphics[width = \columnwidth]{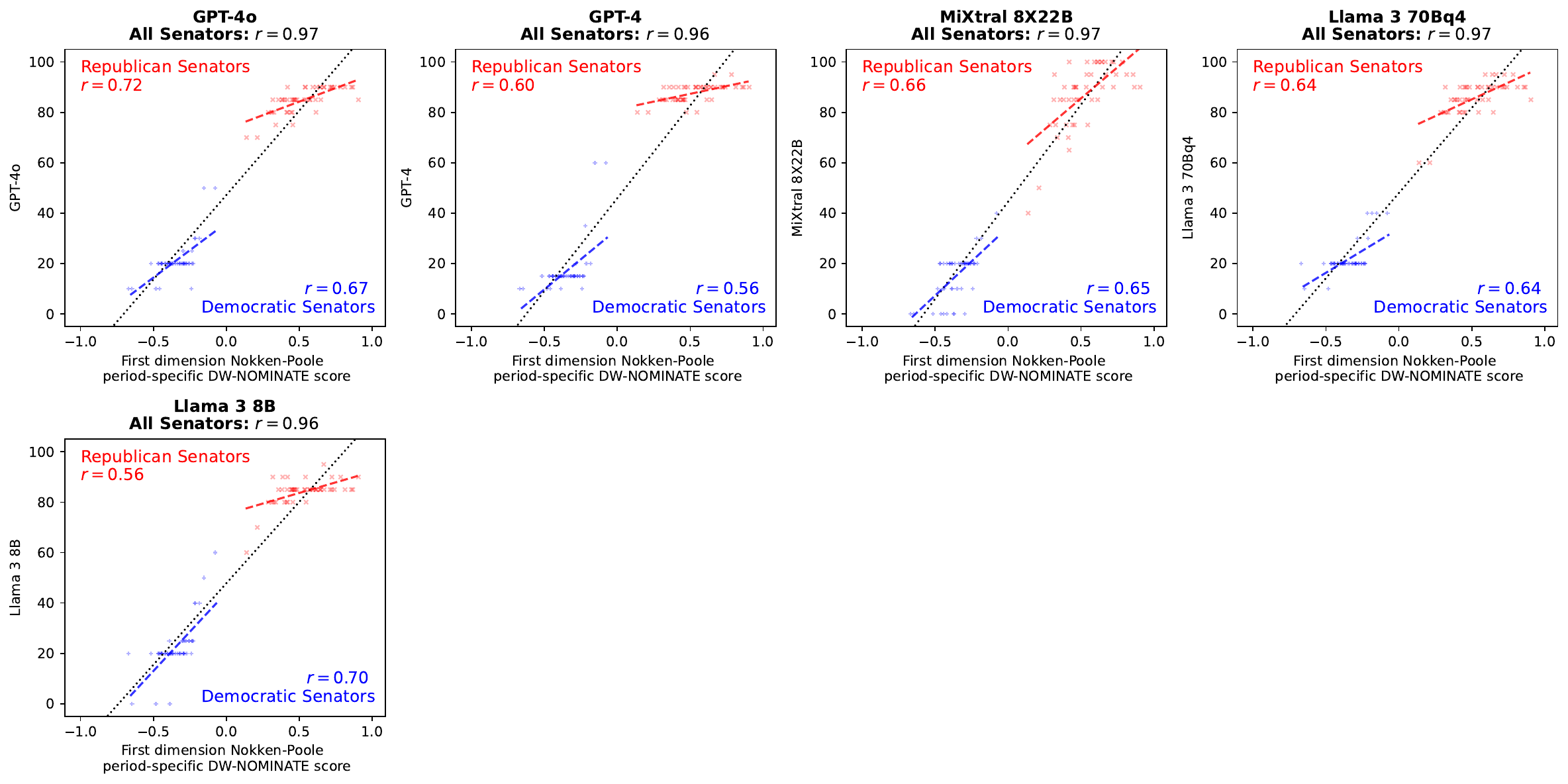}    
    \caption{Replication of Figure~\ref{fig_senator_117_sentence_scaling_Nokken_Poole} by submitting sets of tweets a single prompt. Each dot represents a senator (`{\sf+}': Democrats, `{\sf x}': Republicans, `{\sf o}': others). The Y-axis reports, for each senator, the position estimates of senators based on random samples of 100 of their tweets published during the Congress session. The X-axis reports the first dimension Nokken-Poole period-specific DW-NOMINATE score. 
    }    \label{fig_senator_117_sets_of_tweets_Nokken_Poole}
\end{figure}

\FloatBarrier

\newpage

\section{Additional Results - British Party Manifestos}

\subsection{Performance of position estimates obtained by including descriptions of policy dimensions in the prompts submitted to the LLMs.}

\begin{figure}[htbp]
    \centering
    \includegraphics[width = \columnwidth]{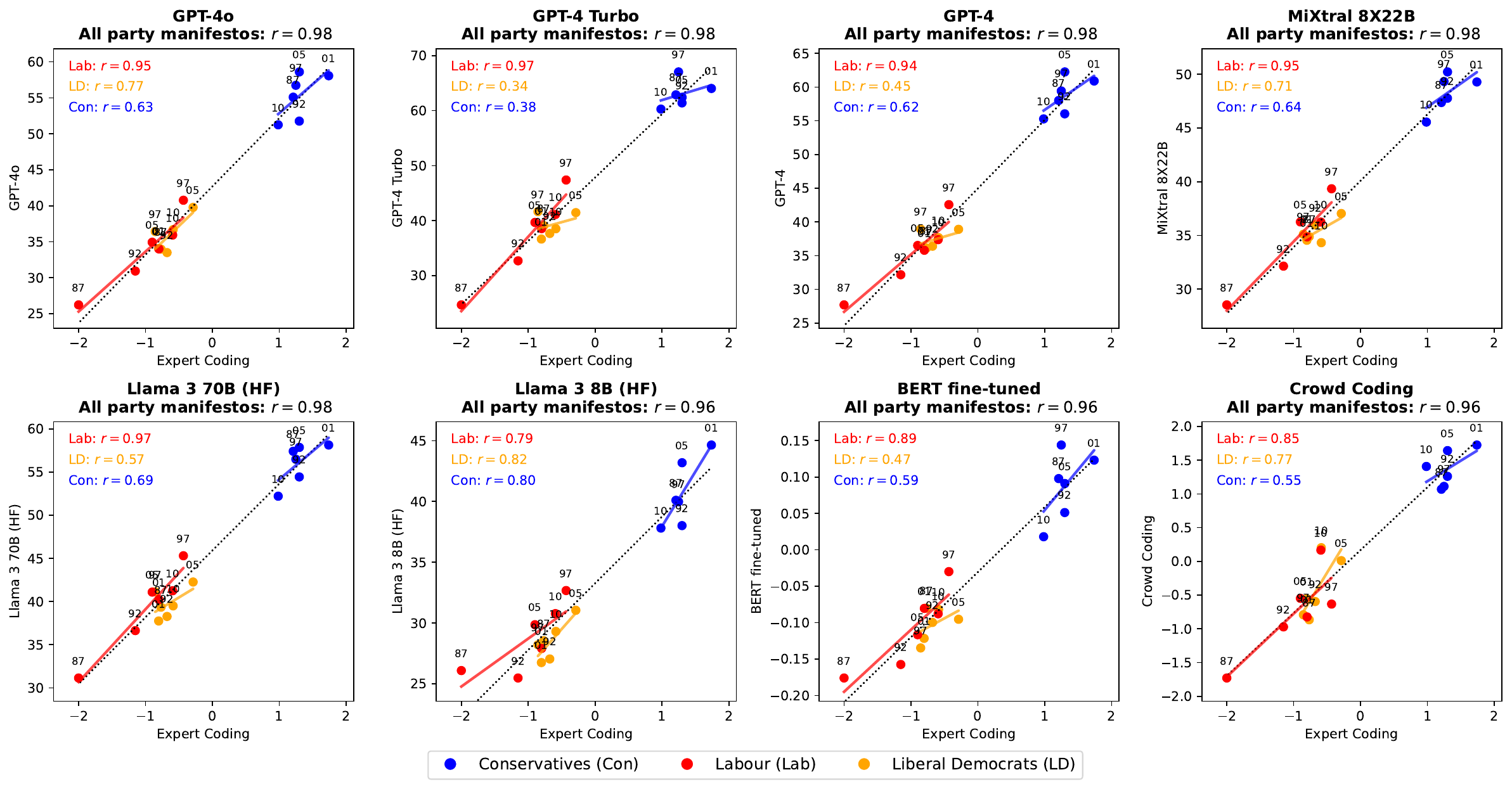}  
    \caption{Positioning British party manifestos on the Economic policy dimension (left to right wing scale, partial replication of figure~\ref{fig_result_UK_manifestos_economic}). A description of the policy dimension and of its extremities as similar as possible to the instructions given to crowdworkers by \cite{benoit2016crowd} was included in the prompts submitted to the LLMs (see Appendix~\ref{app: British party manifestos - scaling with explanations about the policy dimension}).
   \label{fig_result_UK_manifestos_economic_dim_desc}
    }    
    \vspace{0.1cm}

\end{figure}

 \begin{figure}
    \centering
     \includegraphics[width = \columnwidth]{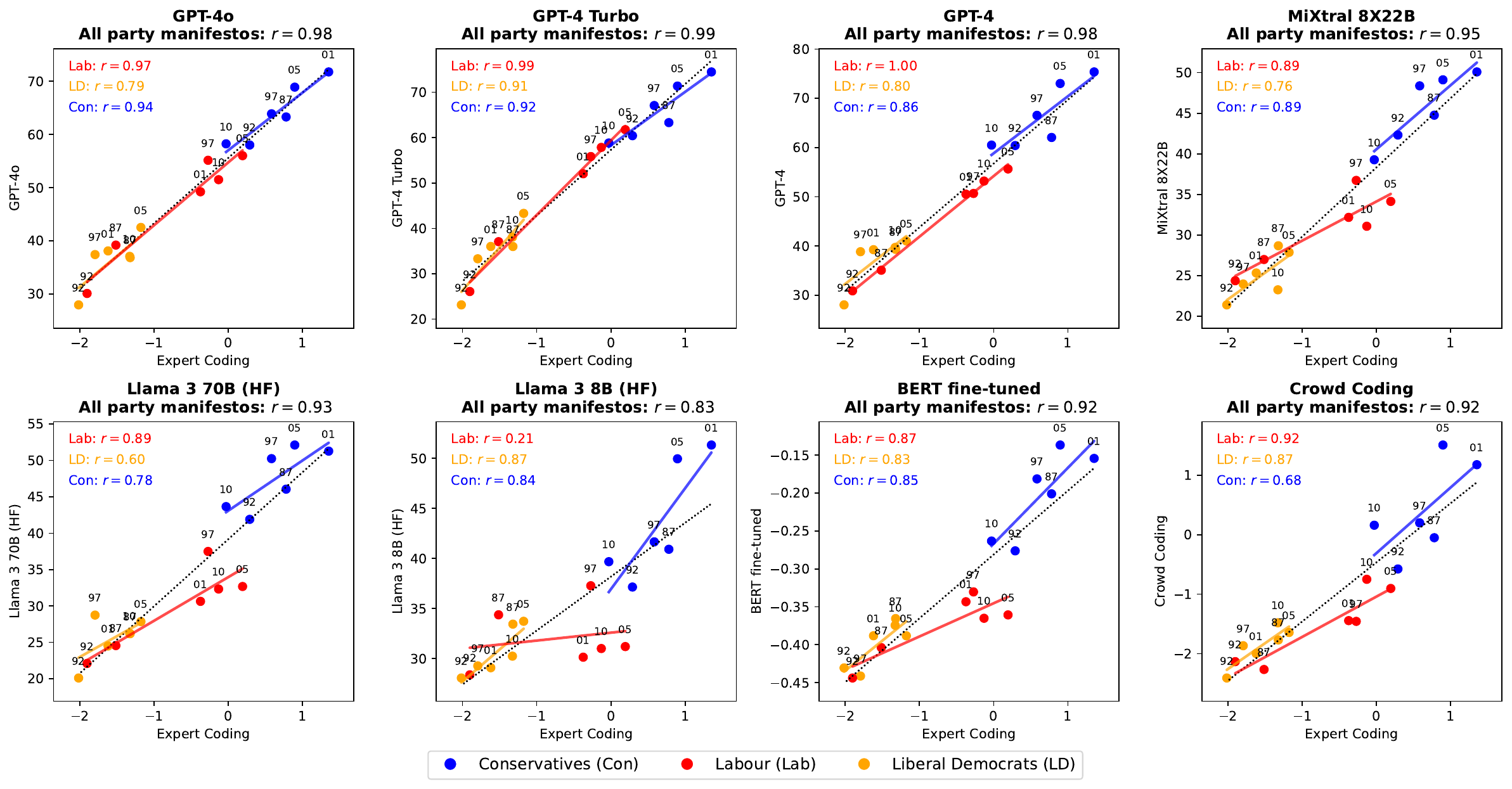}
    
    \caption{Positioning British party manifestos on the Social policy dimension (liberal to conservative scale, partial replication of figure~\ref{fig_result_UK_manifestos_social}). A description of the policy dimension and of its extremities as similar as possible to the instructions given to crowdworkers by \cite{benoit2016crowd} was included in the prompts submitted to the LLMs (see Appendix~\ref{app: British party manifestos - scaling with explanations about the policy dimension}). 
    \label{fig_result_UK_manifestos_social_dim_desc}
    }   
    \vspace{0.1cm}

\end{figure}

\FloatBarrier

\newpage

\subsection{Performance of position estimates obtained by submitting the full text of manifestos in a single prompt to the LLMs.}
\label{app_manifestos_one_prompt}
In the body of the article, we report the results obtained by submitting each sentence of the party manifestos to the LLMs, mimicking the approach adopted by \cite{benoit2016crowd} with the crowds of experts and workers. This method is somewhat slow (it takes about one or two seconds to obtain a position estimate for each sentence, and sometimes longer, for large LLMs executed locally on our laptop). Therefore, we also experimented with an approach that consists in submitting the full text of the manifesto in a single prompt. Among the LLMs we have used, the only ones that have a context window long enough to accept the manifestos in one prompt are GPT-4o, GPT-4-Turbo, and MiXtral 8X22B.

We used a prompt similar to that used for obtaining position estimates of sentences, slightly adapted to the fact that we did not ask for the position of a sentence.
\begin{displayquote}\small
    You will be provided with a text from a party manifesto. 
    
    $\ll$  Text from the party manifesto $\gg$ 
    
    Where does this text stand on the `left' to `right' wing scale, in terms of economic policy? Provide your response as a score between 0 and 100 where 0 means `Extremely left' and 100 means `Extremely right'. To compute the score, start by first identifying the parts of the text that are about economic policy. Then, compute the score based on these parts of text. If the text does not refer to economic policy, return "NA". You will only respond with a JSON object with the key Score. Do not provide explanations.
\end{displayquote}

\begin{figure}[htbp]
    \centering
    \includegraphics[width = \columnwidth]{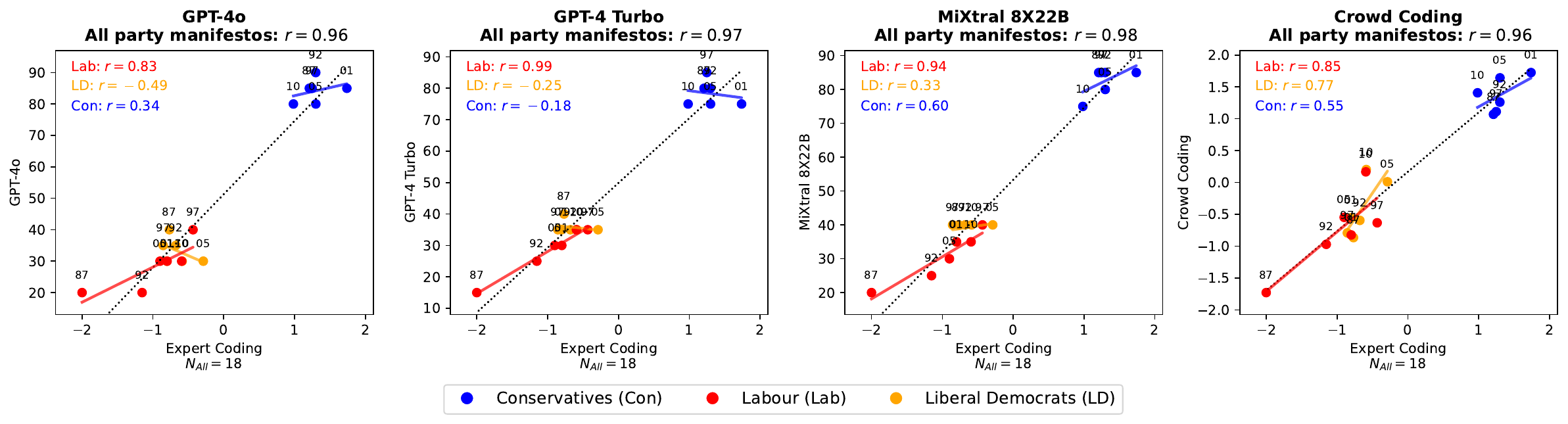}  

    \caption{Positioning British party manifestos on the Economic policy dimension (left to right wing scale) in a single prompt: Pearson correlations between the benchmark (expert survey placement analyzed in \cite{benoit2016crowd}) and position estimates obtained with LLMs or crowdsourced position estimates reported in \cite{benoit2016crowd}. \label{fig_result_UK_manifestos_economic_one_prompt}
    }    
    \vspace{0.1cm}

\end{figure}

\begin{figure}[htbp]
    \centering
     \includegraphics[width = \columnwidth]{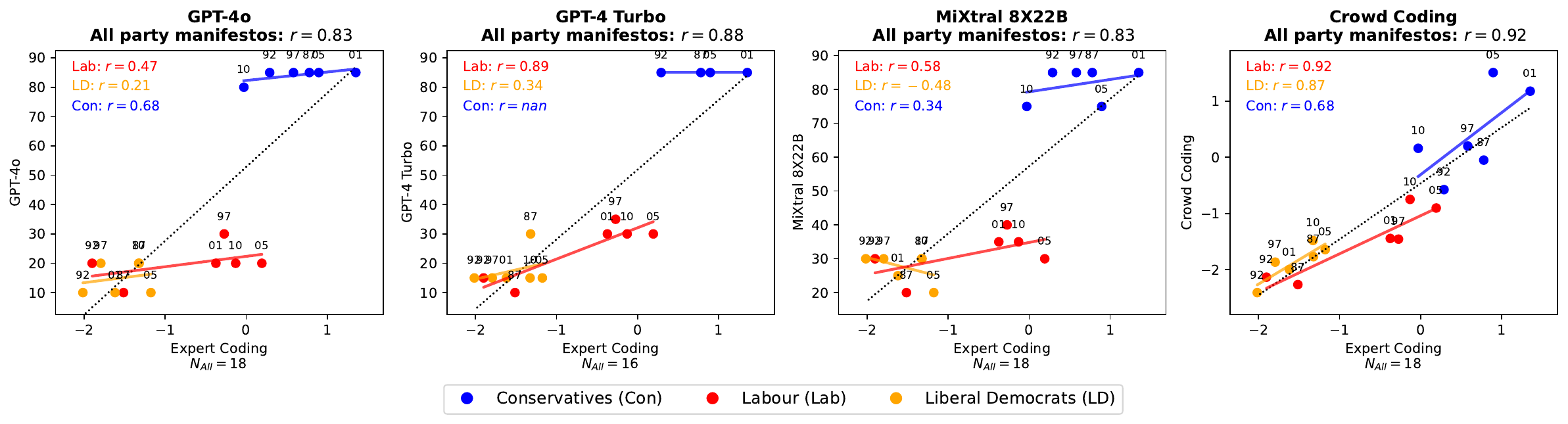}
    \caption{Positioning British party manifestos on the Social policy dimension (liberal to conservative scale) in a single prompt: Pearson correlations between the benchmark (expert survey placement analyzed in \cite{benoit2016crowd}) and position estimates obtained with LLMs or crowdsourced position estimates reported in \cite{benoit2016crowd}. 
    }    \label{fig_result_UK_manifestos_social_one_prompt}
    \vspace{0.1cm}

\end{figure}

The results are reported in figures~\ref{fig_result_UK_manifestos_economic_one_prompt} and~\ref{fig_result_UK_manifestos_social_one_prompt}. In terms of overall correlations, the results are similar to those obtained with the sentence-by-sentence scaling for economic policy, and not as good for social policy. The within-party correlations are not as good as with the sentence-by-sentence approach. We conjecture that this performance difference could come from the similarity between the LLM-based and human-based approaches: when both approaches rely on sentence-by-sentence coding, the results obtained by both approaches correlate more highly than if the two approaches code text of different lengths. This suggests that using LLMs in a way that mimics the procedure used with human coders might lead to a higher match with the position estimates based on human coding. Further studying this conjecture is an interesting avenue for future research.

It is also worth noting that, even though GPT-4o and MiXtral 8X22B returned position estimates for all party manifestos, GPT-4 Turbo failed to do so for two manifestos (for social policy). In ancillary analyses, we observed this failure to return position estimates for long texts in other settings and with other models as well.

\FloatBarrier

\FloatBarrier

\newpage

\section{Additional Results - EU legislative speeches in 10 Languages}
\label{App_Additional Results - EU legislative speeches in 10 Languages}
\subsection{Result summary}

\begin{figure}[htbp]
    \centering
    \hspace{.02\columnwidth}

    \raisebox{0.35\height}{\centering
    \resizebox{0.5\textwidth}{!}{   \begin{tabular}{llllllr}
\toprule
crowd & English & German & Greek & Italian & Polish & Spanish \\
crowd &  &  &  &  &  &  \\
\midrule
English & 1.00 & 0.93 & 0.94 & 0.95 & 0.96 & 0.94 \\
German &  & 1.00 & 0.94 & 0.92 & 0.92 & 0.92 \\
Greek &  &  & 1.00 & 0.92 & 0.94 & 0.95 \\
Italian &  &  &  & 1.00 & 0.96 & 0.94 \\
Polish &  &  &  &  & 1.00 & 0.94 \\
Spanish &  &  &  &  &  & 1.00 \\
\bottomrule
\end{tabular}

  }}
    \hspace{.13\columnwidth}
    \raisebox{-0.5\height}{\includegraphics[width = .33\columnwidth]{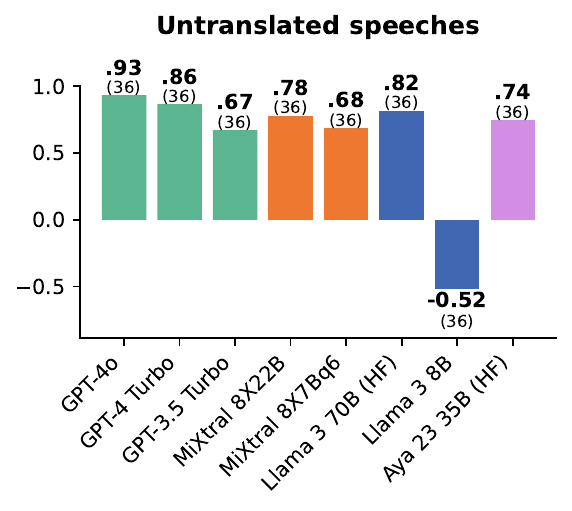}}
    \includegraphics[width = \columnwidth]{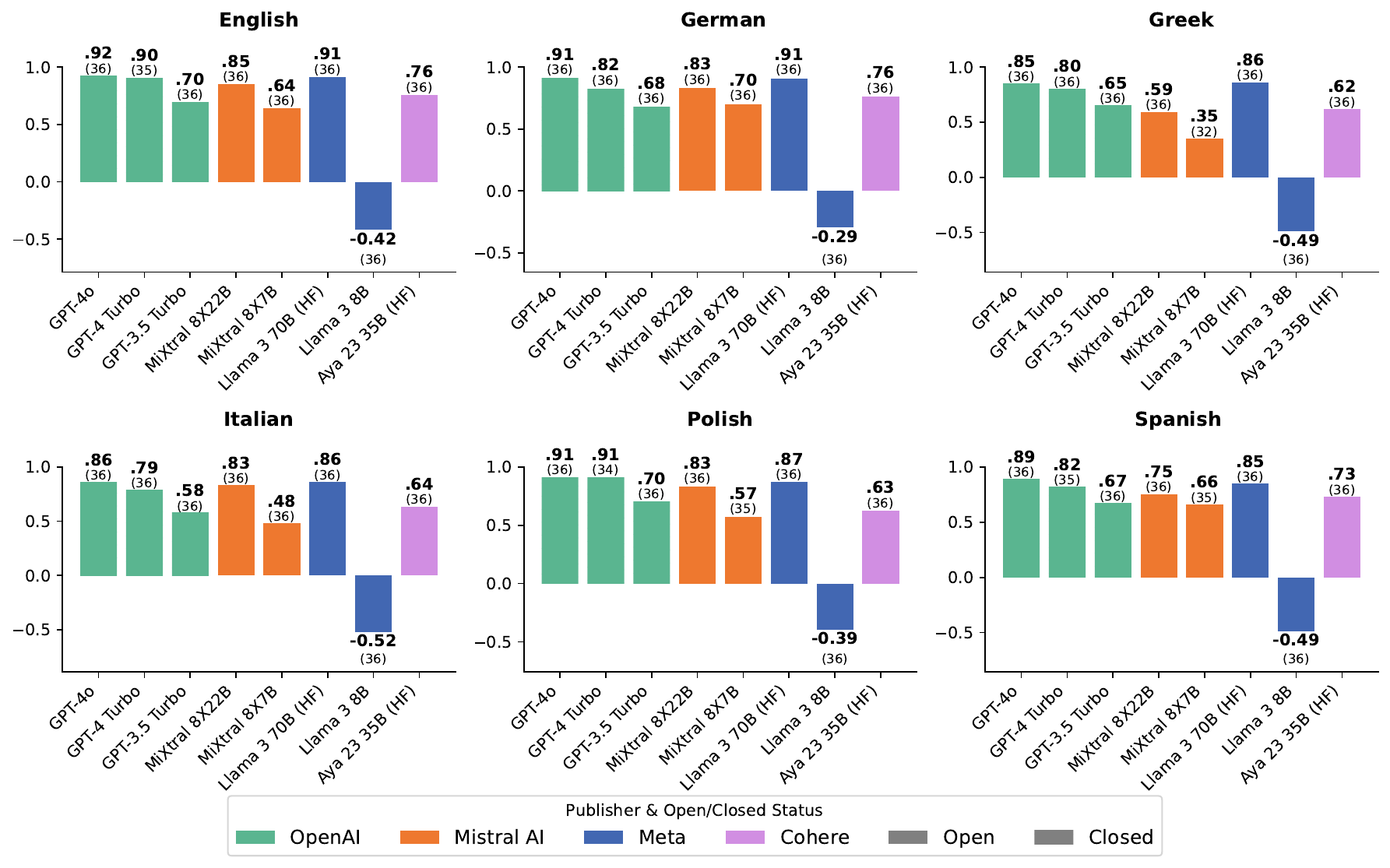}    
    \caption{Positioning EU legislative speeches in 10 languages on the `anti-subsidy’ to ‘pro-subsidy’ dimension: Pearson correlations between the benchmark (crowdsourced position estimates reported in \cite{benoit2016crowd}). The numbers in parentheses indicate the number of documents for which a position was obtained. As a performance benchmark, we report the pairwise correlations between crowdsourced estimates obtained from the translations of speeches in the different languages (upper left panel). These correlations are of similar magnitude. 
    }    \label{fig_result_EU_speeches_appendix}
    \vspace{0.1cm}
\end{figure}

\FloatBarrier

\newpage 

\subsection{Scatter plots for scaling of translations of speeches by LLMs and crowd workers}
\label{app_EU_speeches_translations}

\begin{figure}[htbp]
    \centering
    \includegraphics[width = .9\columnwidth]{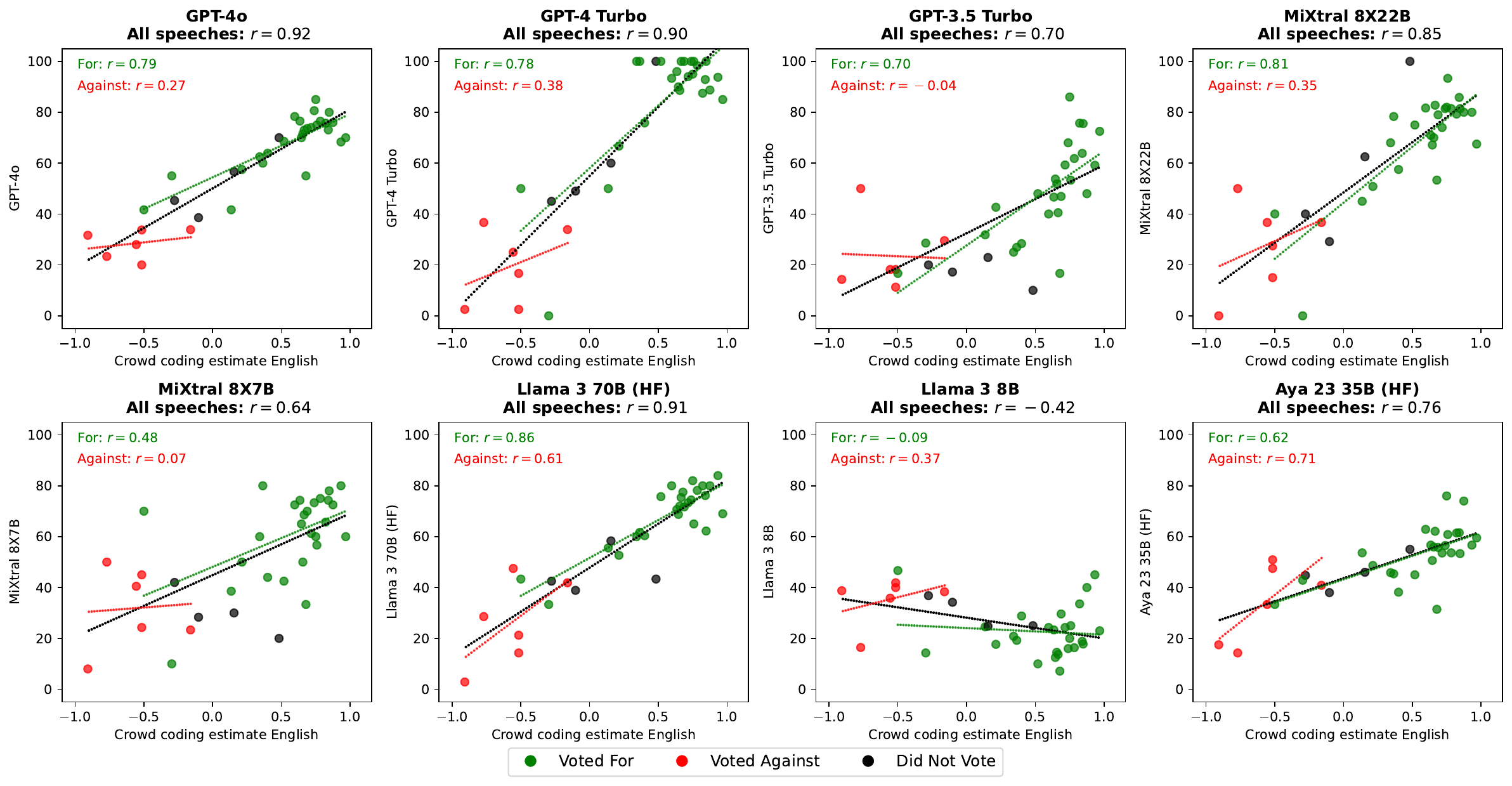}
    \caption{Positioning EU legislative speeches in 10 languages on the `anti-subsidy’ to ‘pro-subsidy’ dimension. Positioning using the official translation in {\bf English}.
    }    \label{fig_EU_speeches_results_EN}
\end{figure}

 \begin{figure}
    \centering
    \includegraphics[width = .9\columnwidth]{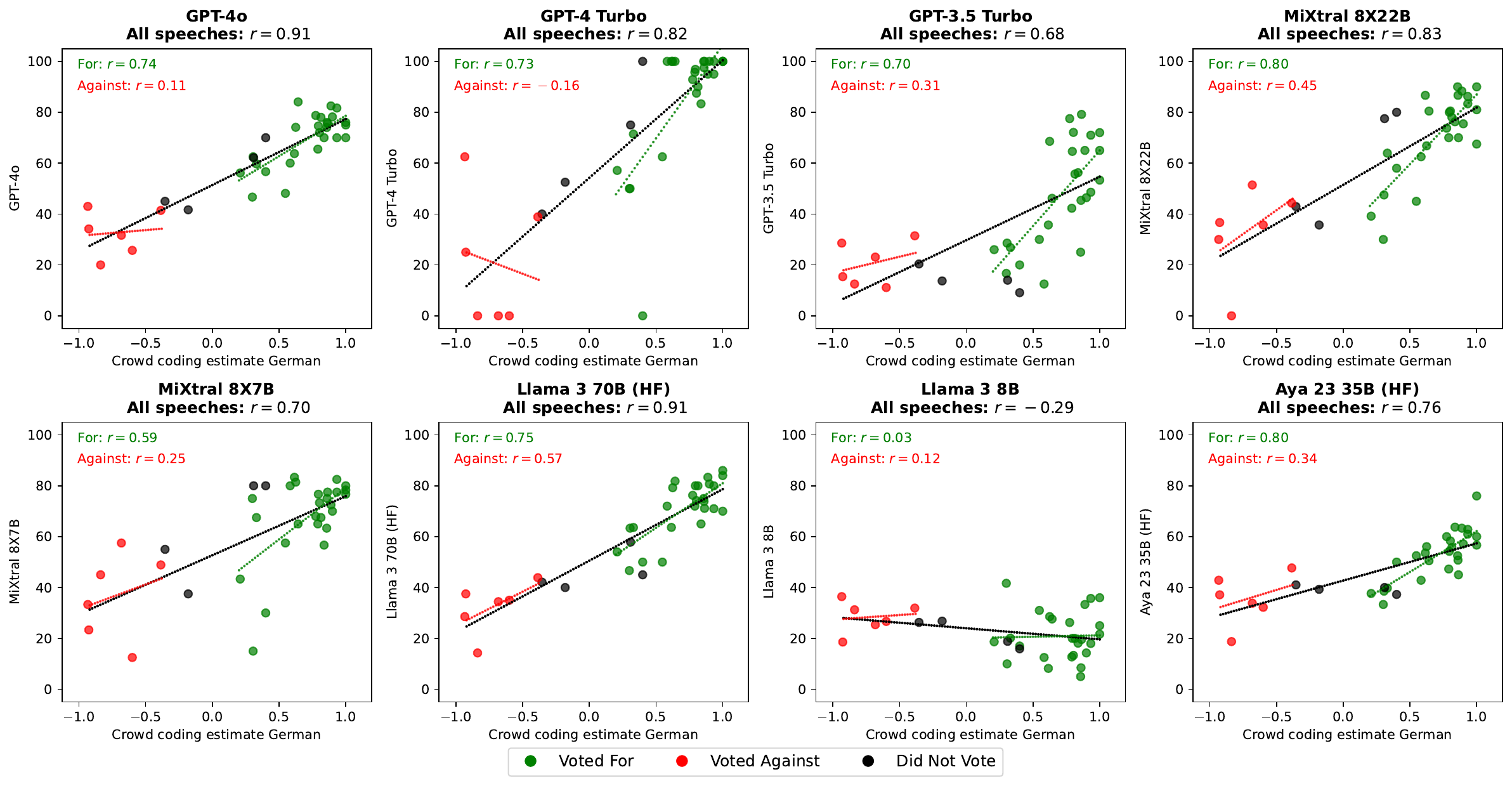}
    \caption{Positioning EU legislative speeches in 10 languages on the `anti-subsidy’ to ‘pro-subsidy’ dimension. Positioning using the official translation in {\bf German}.
    }    
\end{figure}

 \begin{figure}[htbp]
    \centering
    \includegraphics[width = .9\columnwidth]{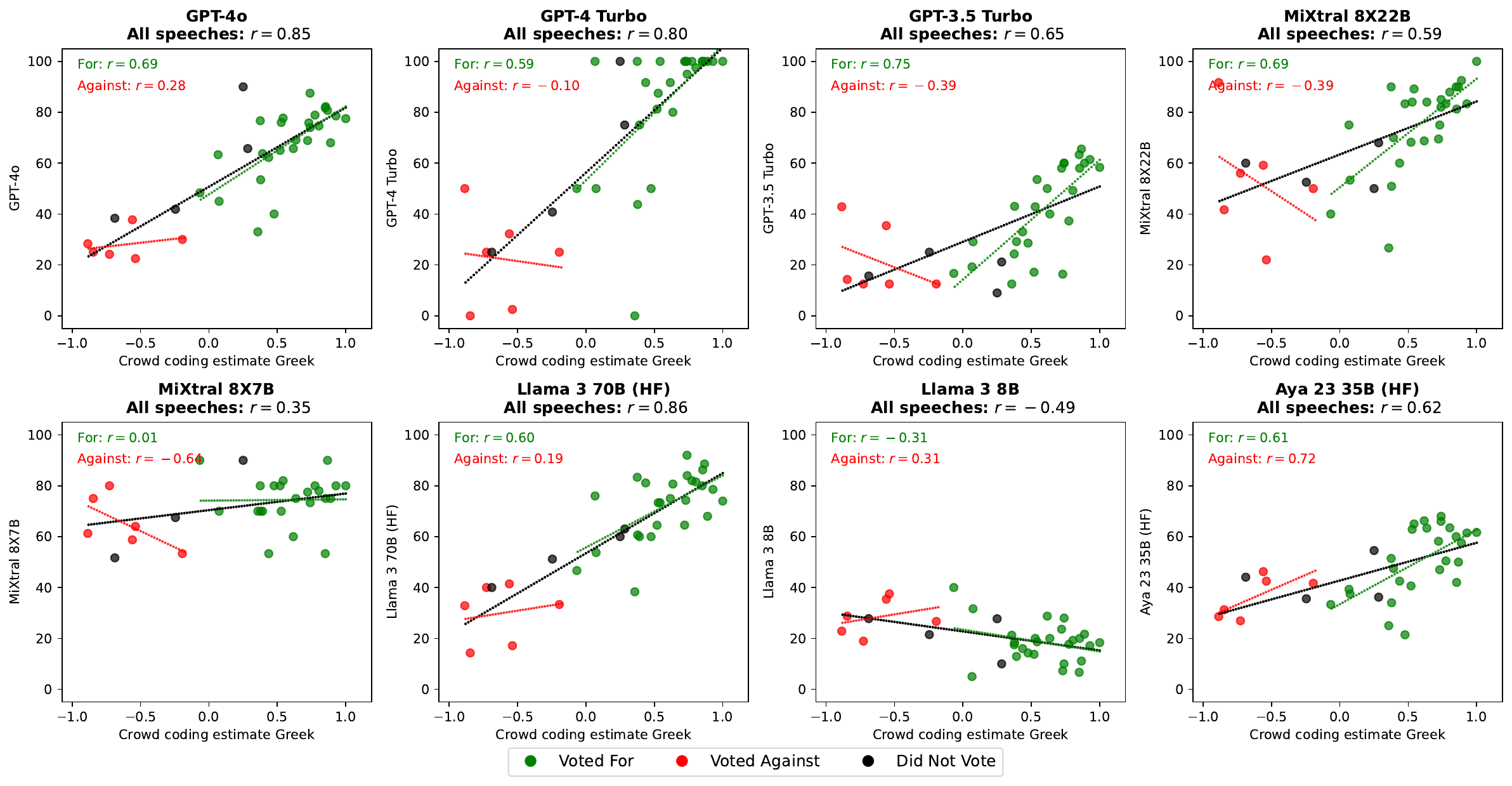}
    \caption{Positioning EU legislative speeches in 10 languages on the `anti-subsidy’ to ‘pro-subsidy’ dimension. Positioning using the official translation in {\bf Greek}.
    }    
\end{figure}

 \begin{figure}[htbp]
    \centering
    \includegraphics[width = .9\columnwidth]{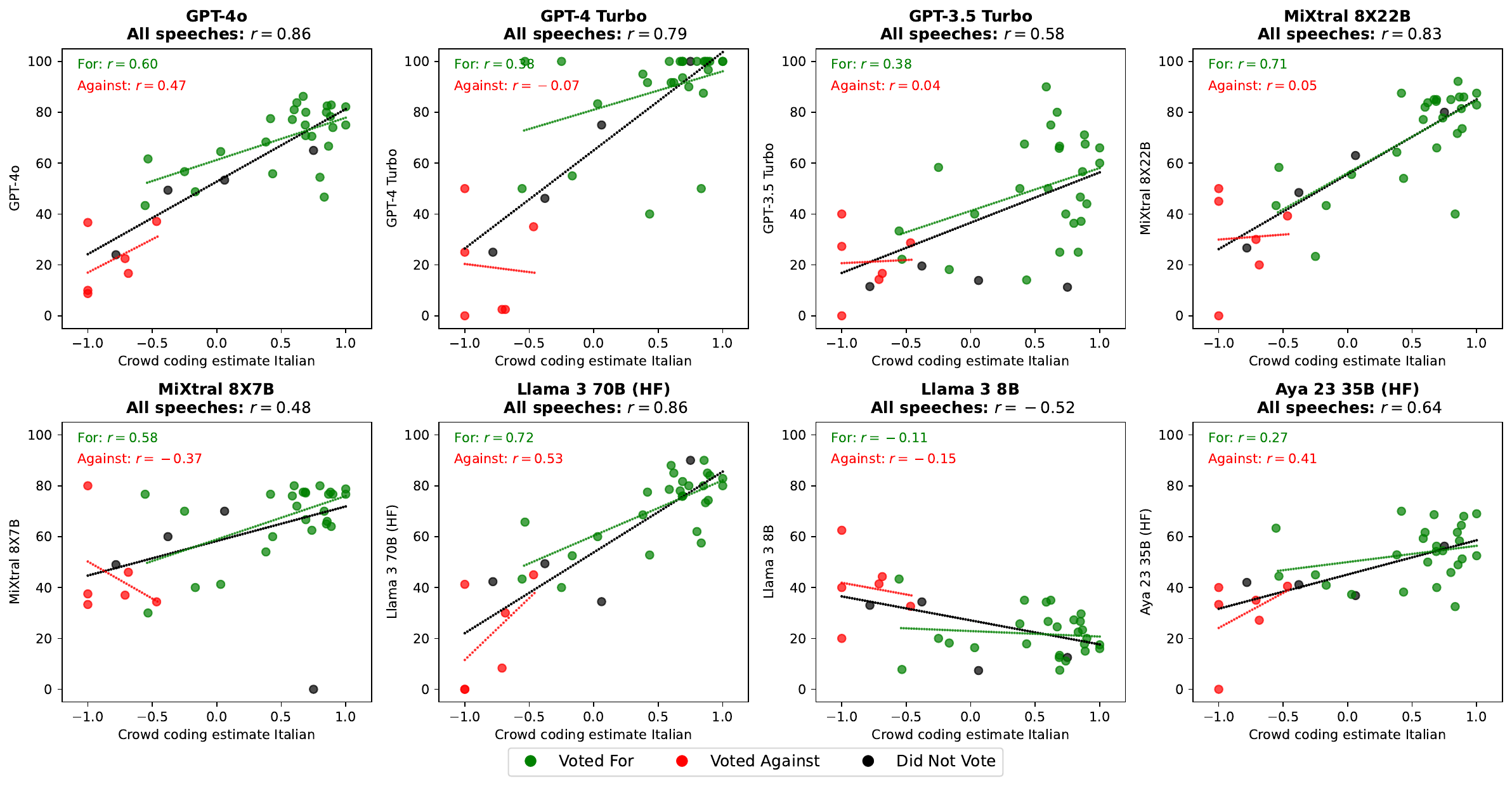}
    \caption{Positioning EU legislative speeches in 10 languages on the `anti-subsidy’ to ‘pro-subsidy’ dimension. Positioning using the official translation in {\bf Italian}.
    }    
\end{figure}

 \begin{figure}[htbp]
    \centering
    \includegraphics[width = .9\columnwidth]{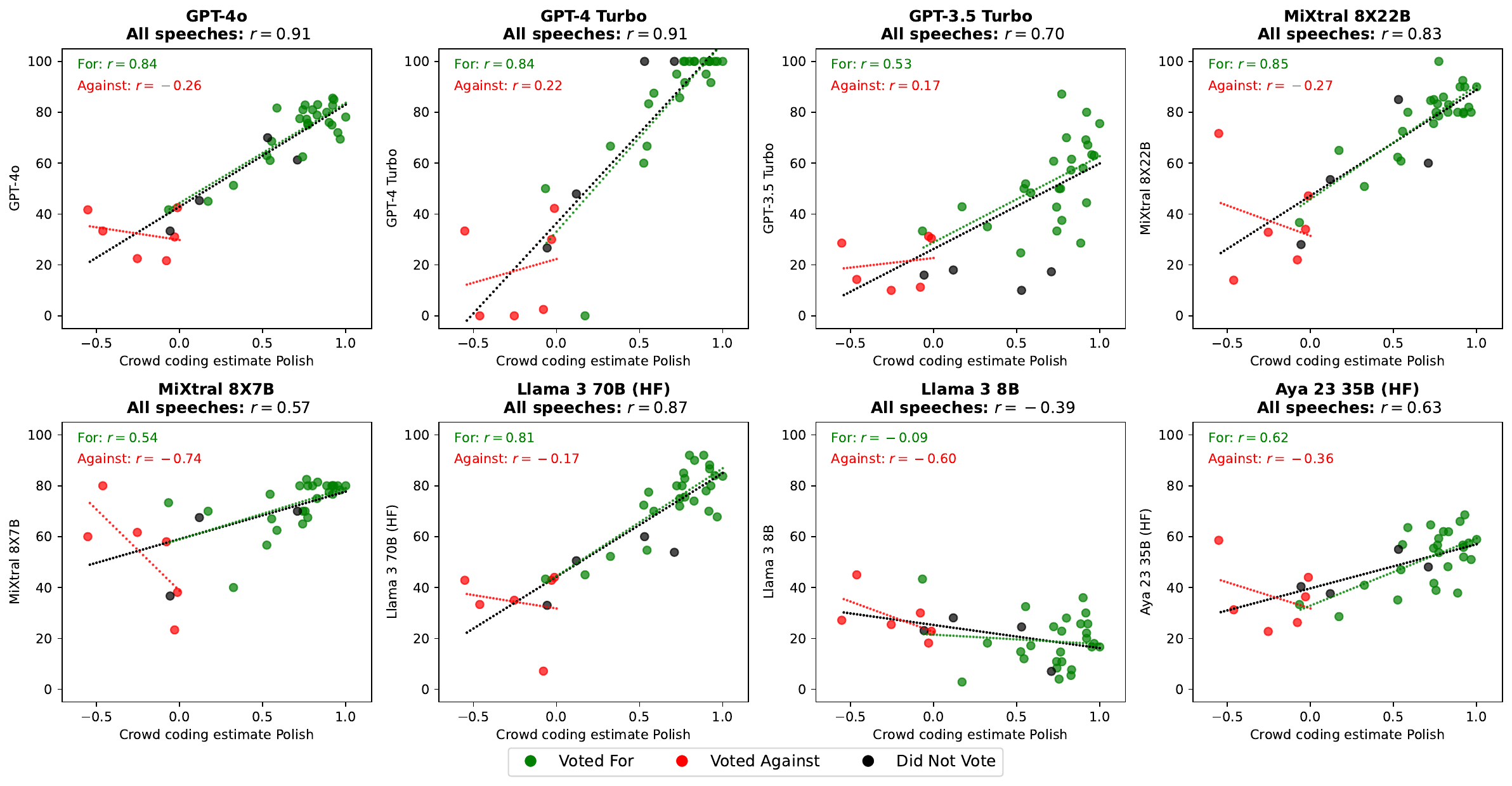}
    \caption{Positioning EU legislative speeches in 10 languages on the `anti-subsidy’ to ‘pro-subsidy’ dimension.  Positioning using the official translation in {\bf Polish}.
    }    
\end{figure}

 \begin{figure}[htbp]
    \centering
    \includegraphics[width = .9\columnwidth]{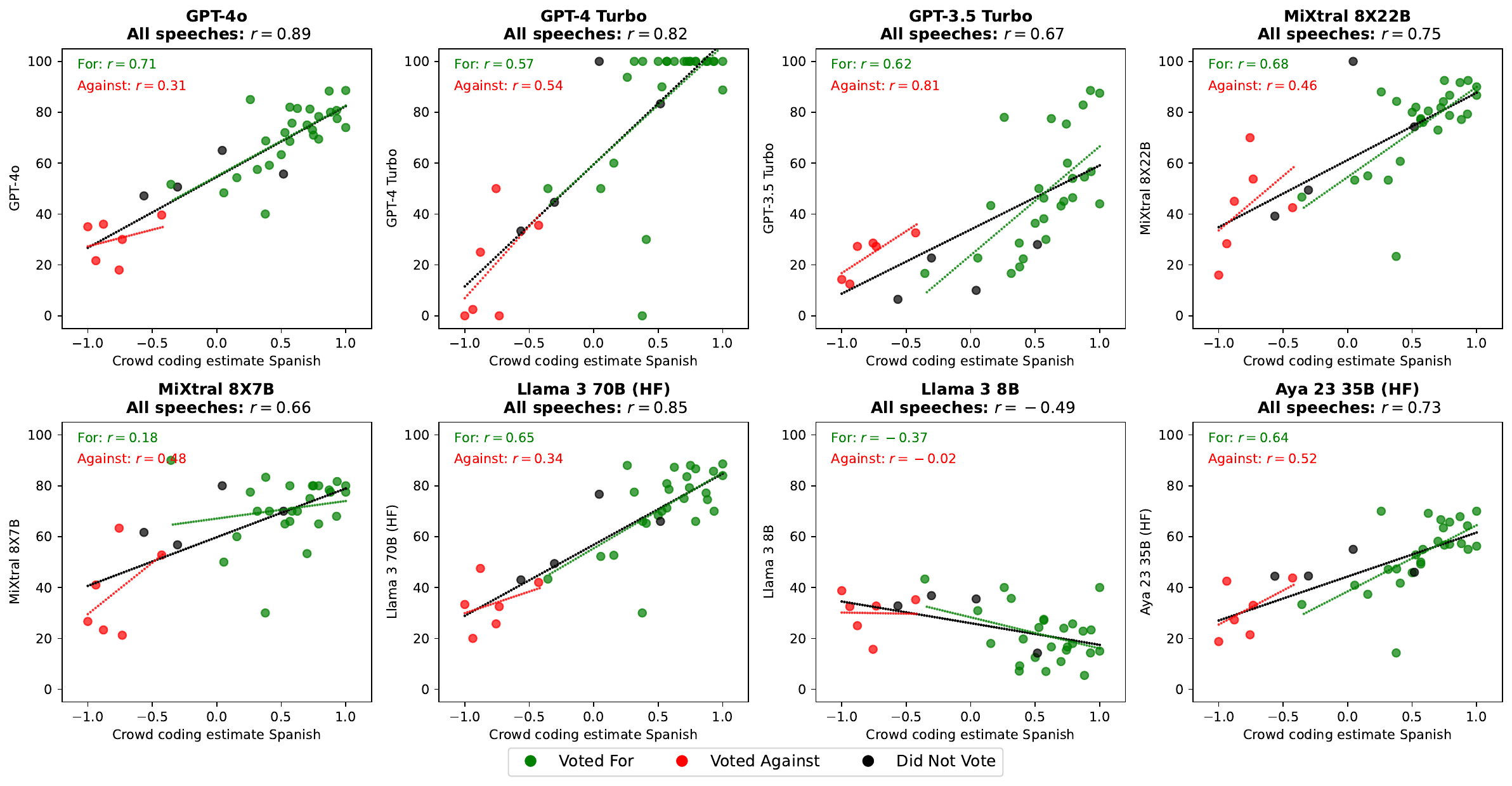}
    \caption{Positioning EU legislative speeches in 10 languages on the `anti-subsidy’ to ‘pro-subsidy’ dimension.  Positioning using the official translation in {\bf Spanish}.
    }    
\end{figure}

\vfill


\end{document}